%% file: arxiv.tex
\documentclass[10pt,journal,compsoc]{IEEEtran}

\usepackage[nocompress]{cite}

\usepackage[margin=0pt,font=small,labelfont=bf,labelsep=endash,tableposition=top]{caption}
\usepackage{epsfig}
\usepackage{graphicx}
\usepackage{amsmath}
\usepackage{amssymb}

\usepackage[accsupp]{axessibility}  %

\usepackage{graphicx}
\usepackage{amsmath}
\usepackage{amssymb}
\usepackage{booktabs}

\usepackage[pagebackref,breaklinks,colorlinks]{hyperref}

\def\etal{{\em et al.}}

\usepackage{multirow}

\usepackage{enumitem}
\usepackage{tabularx}
\usepackage{multirow}
\usepackage{stfloats}
\usepackage{threeparttable}
\usepackage[usenames,dvipsnames]{xcolor}

\usepackage{bm}

\def\etal{{\em et al.}}

\newlength\savedwidth

\definecolor{darkergreen}{RGB}{21, 152, 56}
\definecolor{red2}{RGB}{252, 54, 65}

\definecolor{blacktext}{RGB}{0, 0, 0}
\newcommand\redp[1]{\textcolor{blacktext}{#1}}
\newcommand\greenp[1]{\textcolor{blacktext}{#1}}

\usepackage{xcolor}
\definecolor{yl_color}{RGB}{128, 255, 0}

\definecolor{blue2}{RGB}{20, 54, 254}

\usepackage{bbding}

\usepackage[american]{babel}
\usepackage{microtype}
\usepackage{cleveref}

\usepackage{subcaption}

\begin{document}

\title{Turning a CLIP Model into a Scene Text Spotter}

\author{
Wenwen Yu,
Yuliang Liu*, IEEE Member,
Xingkui Zhu,
Haoyu Cao,
Xing Sun, \\
Xiang Bai, IEEE Senior Member
\IEEEcompsocitemizethanks{
\IEEEcompsocthanksitem W. Yu, Y. Liu, X. Zhu, and X. Bai are with the School of Artificial Intelligence and Automation, Huazhong University of Science and Technology, Wuhan, 430074, China (email: \{wenwenyu, ylliu, adlith, xbai\}@hust.edu.cn).
\IEEEcompsocthanksitem H. Cao and X. Sun are with the Tencent YouTu Lab, Hefei, 230000, China (email: rechycao@tencent.com, winfred.sun@gmail.com).
}
\thanks{Corresponding author: Yuliang Liu.}
}

\IEEEtitleabstractindextext{%
\begin{abstract}

    We exploit the potential of the large-scale Contrastive Language-Image Pretraining (CLIP) model to enhance scene text detection and spotting tasks, transforming it into a robust backbone, FastTCM-CR50. This backbone utilizes visual prompt learning and cross-attention in CLIP to extract image and text-based prior knowledge. Using predefined and learnable prompts, FastTCM-CR50 introduces an instance-language matching process to enhance the synergy between image and text embeddings, thereby refining text regions. Our Bimodal Similarity Matching (BSM) module facilitates dynamic language prompt generation, enabling offline computations and improving performance. FastTCM-CR50 offers several advantages: 1) It can enhance existing text detectors and spotters, improving performance by an average of 1.7\% and 1.5\%, respectively. 2) It outperforms the previous TCM-CR50 backbone, yielding an average improvement of 0.2\% and 0.56\% in text detection and spotting tasks, along with a 48.5\% increase in inference speed. 3) It showcases robust few-shot training capabilities. Utilizing only 10\% of the supervised data, FastTCM-CR50 improves performance by an average of 26.5\% and 5.5\% for text detection and spotting tasks, respectively. 4) It consistently enhances performance on out-of-distribution text detection and spotting datasets, particularly the NightTime-ArT subset from ICDAR2019-ArT and the DOTA dataset for oriented object detection. The code is available at \url{https://github.com/wenwenyu/TCM}.
\end{abstract}

}

\maketitle

\

\IEEEraisesectionheading{\section{Introduction}
\label{sec:introduction}}
  \IEEEPARstart{S}{c}ene text spotting, aiming at the localization and recognition of text instances within natural images, has remained at the forefront due to its diverse practical applications, which include online education, office automation, automatic driving, and instant translation. The evolution of fully-supervised deep learning technologies has spearheaded substantial advancements within scene text spotting. Yet, these supervised methodologies, are heavily reliant on detailed and extensive annotations, indicating a potential limitation when facing scenarios with divergent data distributions. How to improve the performance of text spotting techniques under circumstances of sparse annotated data or when shifting between different domains - commonly referred to as few-shot training and generalization ability - is increasingly gaining attention.

    \begin{figure}[tbp]
        \centering
        \includegraphics[width=0.48\textwidth]{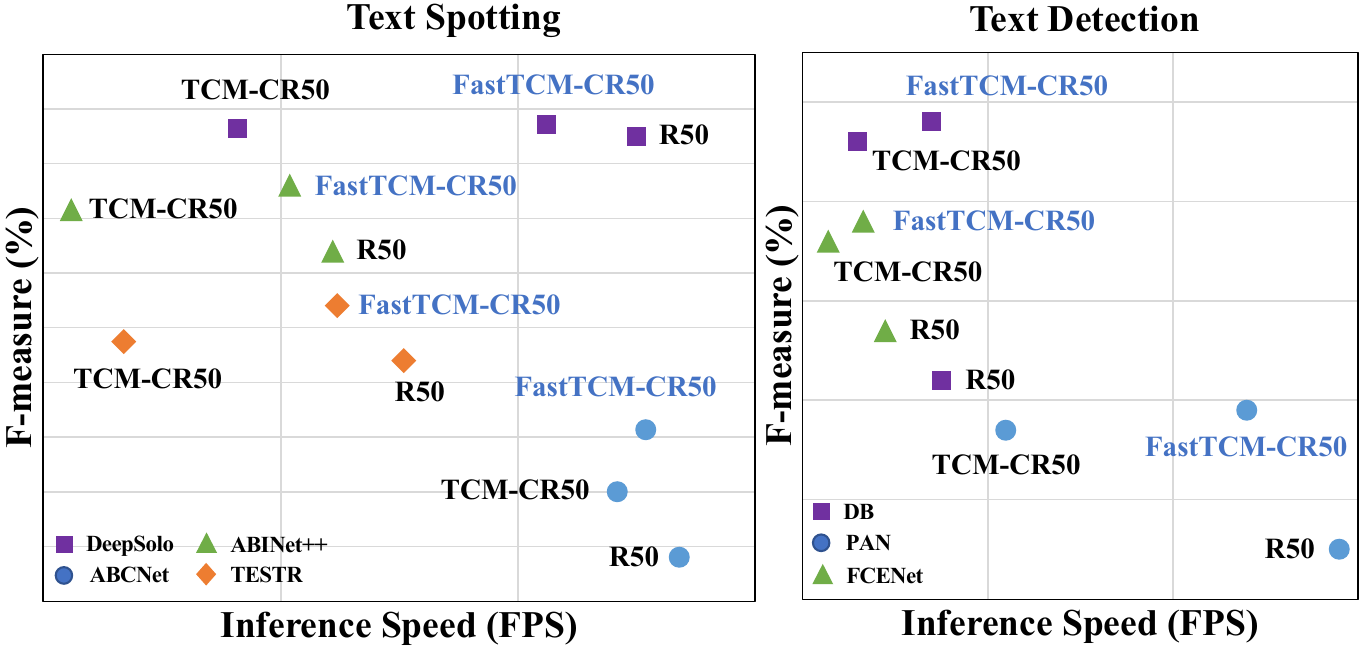}
        \caption{Comparison of F-measure and FPS among different backbones on text spotting and text detection methods. FastTCM-CR50 achieves higher performance while significantly improving the inference speed compared to TCM-CR50. The text spotting F-measure is calculated across IC15, Total-Text, and CTW1500. The text detection F-measure is calculated across IC15, TD, and CTW1500, both averaged across the datasets.
        }
        \label{fig:intro_speed_compare}
    \end{figure}

  In the past decade, utilizing the backbones such as VGG16 and ResNet-50 from ImageNet and MSCOCO to acquire better initialization and generalization ability for scene text detection and spotting are commonly adopted as a basic setting. Recently, developments in leveraging pretrained vision and language knowledge, particularly through the large-scale Contrastive Language-Image Pretraining (CLIP) model~\cite{Radford2021LearningTV}, have shown promising results in a wide range of downstream tasks. These include but are not limited to image classification~\cite{Zhou2022ConditionalPL}, object detection~\cite{Gu2022OpenvocabularyOD}, and semantic segmentation~\cite{Rao2022DenseCLIPLD, Xu2021ASB}. In the realm of text spotting, where scene text often provides rich visual and character information, the potential of the CLIP model is particularly evident. How to excavate cross-modal information from visual, semantic, and text knowledge to enhance the performance of text detection and spotting models has gained more and more attention. Song et al.~\cite{Song2022VisionLanguagePF}, for instance, has proposed a fine-grained cross-modality interaction approach, inspired by CLIP, to align unimodal embeddings and improve the learning of representations through pretraining tasks for scene text detection. 
  Wan et al.\cite{Wan2021SelfattentionBT} have brought forth an approach that involves a self-attention based text knowledge mining technique to boost the backbone via image-level text recognition pretraining tasks. Meanwhile, Xue et al.\cite{Xue2022LanguageMA} have introduced a weakly supervised pretraining method aiming to jointly learn and align visual and partial textual information. The goal is to cultivate effective visual text representations applicable to scene text detection and spotting. 
    
    \begin{figure*}[tbp]
        \centering
        \includegraphics[width=0.97\textwidth]{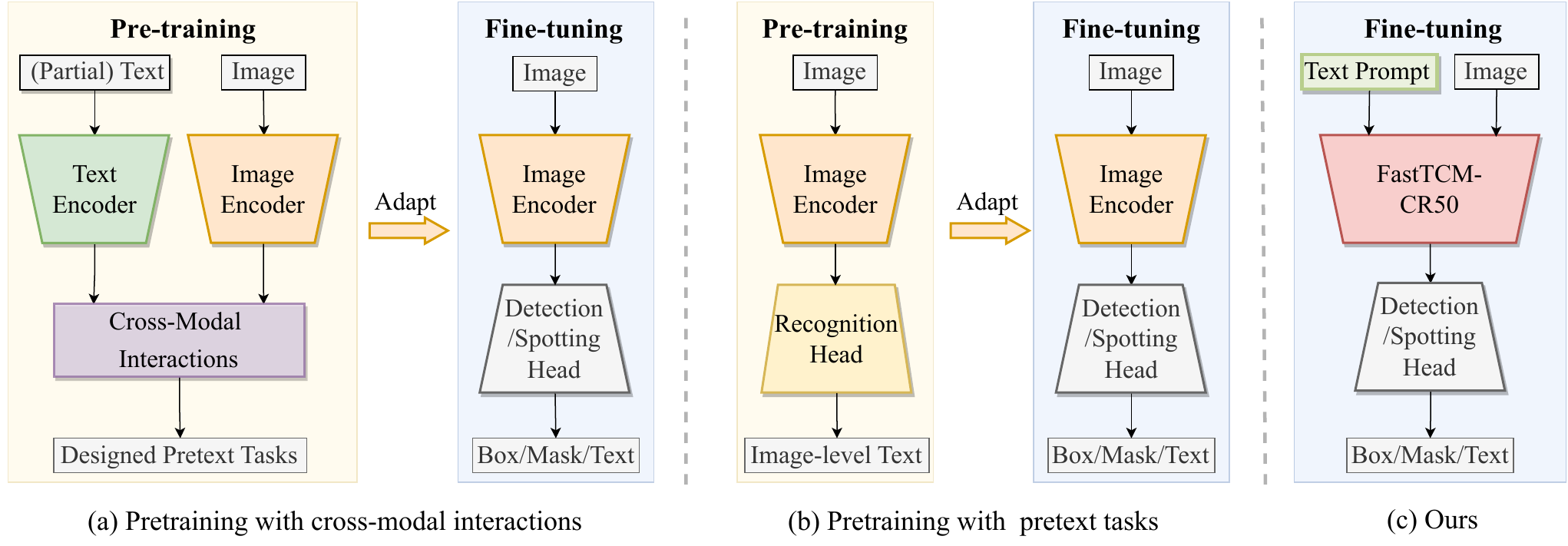}
        \caption{We compare different paradigms of utilizing text knowledge for scene text detection and spotting. Our approach directly delivers an enhanced CLIP backbone, eliminating the need for a pretraining process that relies on specifically designed pretext tasks. CR50 represents for CLIP-ResNet50 model.
        }
        \label{fig:compare_intro}
    \end{figure*}
    
    Contrary to existing approaches illustrated in Fig.~\ref{fig:compare_intro}, our aim is to transform the CLIP model directly into a foundation for text detection and spotting, eliminating the need for pretraining process. However, this is not a straightforward task, as we empirically observe that only solely employing the CLIP model leads to minimal enhancements, and even worse results in aerial object detection, as shown in Sec.~\ref{subsec:rod}. The primary challenge lies in identifying an effective method to leverage visual and semantic prior information specific to each image.

    To this end, we introduce a new backbone specifically designed for scene text detection and spotting tasks, termed as \textbf{FastTCM-CR50}. 
    This model can be conveniently incorporated into existing scene text detection and spotting frameworks to enhance their performance. Central to our approach is a cross-modal interaction mechanism established through visual prompt learning. The mechanism, realized via cross-attention, recovers the locality feature from the CLIP image encoder, thereby capturing fine-grained information for the subsequent matching of text instances with the language, which is particularly useful in responding to coarse text regions. Further, to exploit the prior knowledge from the text encoder for different input images, we utilize an improved language prompt generator built on the learnable query and bimodal similarity matching to obtain global image information. In addition, we have also devised an instance-language matching method to align the image and text embeddings, which aids the image encoder to refine text regions based on cross-modal visual-language priors. The FastTCM-CR50 model can then be directly fine-tuned for the text detection and spotting tasks without requiring a pretext task pretraining, as detailed in Fig.~\ref{fig:compare_intro}(c). Compared to our conference version TCM-CR50~\cite{Yu2023TurningAC}, FastTCM-CR50 introduces Bimodal Similarity Matching (BSM) module as well as the learnable parameters as an implicit image condition that enables and further enhances the CLIP text encoder to perform offline calculations during inference, thereby achieving better results and reducing the inference time, as shown in Fig.~\ref{fig:intro_speed_compare}. 
    
     We summarize the advantages of our method as follows:
     \begin{itemize}
        \item Our proposed FastTCM-CR50 backbone inherently enhances current scene text detectors and spotters, resulting in average performance improvements with numerous baseline methods by 1.7\% and 1.5\% for scene text detection and spotting tasks, respectively. 
        \item Besides, FastTCM-CR50 outperforms the previous text detection and spotting backbone TCM-CR50, delivering an average performance boost of 0.2\% and 0.56\% in text detection and spotting tasks, respectively, along with a notable 48.5\% increase in inference speed.
        \item Demonstrating robust few-shot training capabilities, our new backbone, when trained with only 10\% of the supervised data, exhibits an impressive average performance surge of 26.5\% and 5.5\% for text detection and spotting tasks, respectively.
        \item  In terms of generalization ability, our approach notably surpasses baseline methods by an average of 12.4\% and 14.8\% for domain adaptation tasks for text detection and spotting, respectively. Particularly noteworthy are the significant improvements achieved on the NightTime-ArT subset from ICDAR2019-ArT and the rotated object detection dataset, DOTA-v1.0, illustrating its robust generalization capabilities across diverse task domains.
     \end{itemize}

\section{Related works}
\label{sec:rela}
    \subsection{Scene Text Detection} 
    Scene text detection is a technique that exclusively utilizes bounding box annotations. This method can generally be categorized into two primary types: segmentation-based and regression-based techniques.

    \noindent \textbf{Segmentation-based Methods.}  
    Segmentation-based techniques typically perform operations at the pixel, segment, or contour level, subsequently grouping these into text instances. Notable methods include the Segment Linking (SegLink) by Shi et al.\cite{Shi2017DetectingOT}, using a fully-convolutional neural network for detecting segments and links; the TextSnake by Long et al.\cite{Long2018TextSnakeAF}, an adaptable approach for detecting text of varying shapes; and the Progressive Scale Expansion Network (PSENet) by Li et al.~\cite{Li2019ShapeRT}, generating diverse kernel scales for each text instance. Efficient and accurate systems like the Pixel Aggregation Network (PAN) developed by Wang et al.\cite{Wang2019EfficientAA} have emerged, combining a low computational-cost segmentation head with a learnable post-processing system. Additionally, unique methods such as the Differentiable Binarization (DB) module introduced by Liao et al.\cite{Liao2020RealtimeST} incorporate the binarization step directly into the segmentation network. Meanwhile, the transformer-based architecture proposed by Tang et al.~\cite{Tang2022FewCB} performs detection based on select representative features to decrease computational cost and reduce background interference. These methods underscore the wide range and adaptability of segmentation-based techniques in text detection. Further, Long et al.~\cite{Long2022TowardsEU} introduced an end-to-end model capable of performing unified scene text detection and visual layout analysis simultaneously.
    
    \noindent \textbf{Regression-based Methods.} 
    Regression-based methods view text as a single object and directly regress the bounding boxes of the text instances. Zhang et al.\cite{Zhang2016MultiorientedTD} propose a multi-oriented text detection method utilizing Fully Convolutional Networks, which uses both local and global cues to locate text lines. Liu et al.\cite{Liu2017DeepMP} develop the Deep Matching Prior Network (DMPNet), using quadrilateral sliding windows and a sequential protocol for regression to predict text with a compact quadrangle. He et al.\cite{He2017SingleST, He2017DeepDR} introduced models for text detection that utilize a regional attention mechanism and deep direct regression to predict the text bounding box. Liao et al.\cite{Liao2017TextBoxesAF} create a unified deep neural network for natural image text detection, and Zhou et al.~\cite{Zhou2017EASTAE} designed the EAST model that predicts words or text lines of any orientation and quadrilateral shape in full images. Innovative methods like LOMO by Zhang et al.\cite{Zhang2019LookMT}, and the adaptive text region representation by Wang et al.~\cite{Wang2019ArbitrarySS} have also been developed. Zhu et al.'s FCENet~\cite{zhu2021fourier}, Liu et al.'s MOST~\cite{He2021MOSTAM}, and Zhang et al.'s adaptive boundary proposal network~\cite{Zhang2021AdaptiveBP} further contribute to the field by introducing novel concepts and methodologies. Dai et al.\cite{Dai2021ProgressiveCR} use a progressive contour regression strategy, and Ye et al.'s DPText-DETR\cite{Ye2022DPTextDETRTB} employs explicit point coordinates and an enhanced self-attention module. Zhang et al.~\cite{Zhang2022ArbitraryST} present a unified coarse-to-fine framework for text detection using an iterative boundary transformer.

    \subsection{Scene Text Spotting} 
    Scene text spotting typically employs a unified end-to-end trainable network, blending text detection and text recognition into a cross-modal assisted paradigm. This integrated approach streamlines text detection and recognition into a singular network. It enables simultaneous localization and identification of text within images, capitalizing on the synergistic relationship between text detection and recognition to augment overall performance. Scene text spotting can be bifurcated into two principal categories: regular end-to-end scene text spotting and arbitrarily-shaped end-to-end scene text spotting. Regular end-to-end scene text spotting concentrates on discerning and deciphering text within standard-shaped regions, whereas arbitrarily-shaped end-to-end scene text spotting broadens its scope to manage text in irregular or curved formations.

    \noindent\textbf{Regular End-to-end Scene Text Spotting.}
    Li et al.~\cite{li2017towards} propose one of the earliest end-to-end trainable scene text spotting methods. Their approach effectively merged detection and recognition features using RoI Pooling~\cite{ren2015faster} in a two-stage framework. Originally designed for horizontal and focused text, their method showed significant performance improvements in an enhanced version~\cite{li2019towards}. Busta et al.~\cite{busta2017deep} made contributions to the field with their end-to-end deep text spotter. In further advancements, He et al.~\cite{he2018end} and Liu et al.~\cite{liu2018fots} incorporated anchor-free mechanisms to enhance training and inference speed. They employed different sampling strategies, such as Text-Align-Sampling and RoI-Rotate, respectively, to extract features from quadrilateral detection results.

    \noindent\textbf{Arbitrarily-shaped End-to-end Scene Text Spotting.} 
    Liao et al.\cite{lyu2018mask} introduced Mask TextSpotter which uses Mask R-CNN with character-level supervision to detect and recognize arbitrarily-shaped text. Mask TextSpotterv2\cite{liao2019mask} reduces reliance on character-level annotations. Qin et al.\cite{qin2019towards} employ RoI Masking for attention on arbitrarily-shaped text regions. Feng et al.\cite{feng2019textdragon} utilize RoISlide for handling long text, whereas Wang et al.\cite{Wang2020AllYN} focus on boundary points detection, text rectification, and recognition. CharNet by Xing et al.\cite{xing2019convolutional} also caters to arbitrarily-shaped text spotting. Liao et al.'s Segmentation Proposal Network (SPN)\cite{Liao2020MaskTV} and Liu et al.'s ABCNet~\cite{Liu2020ABCNetRS} are other noteworthy contributions. ABINet++ by Fang et al.~\cite{Fang2022ABINetAB} innovatively uses a vision model and a language model with an iterative correction mechanism. Huang et al.'s SwinTextSpotter~\cite{Huang2022SwinTextSpotterST} uses a transformer encoder for detection and recognition. Approaches based on DETR~\cite{Carion2020EndtoEndOD} and variants~\cite{Zhu2020DeformableDD} for RoI-free scene text spotting have also shown promising results. TESTR~\cite{Zhang2022TextST} uses an encoder and dual decoders, while TTS~\cite{Kittenplon2022TowardsWT} uses a transformer-based approach. SPTS~\cite{Peng2021SPTSST} employs a single point for each instance and uses a Transformer to predict sequences. DeepSolo~\cite{Ye2022DeepSoloLT} allows a single decoder to perform text detection and recognition.

    \subsection{Cross-modal Pretraining Methods} 
    Cross-modal assisted methods leverage a rich blend of cross-modal information including visual, semantic, and text data to amplify the performance for scene text detection and spotting tasks. Wan et al.~\cite{Wan2021SelfattentionBT}, for instance, implemented image-level text recognition pretraining tasks to fortify the backbone using their proposed self-attention-based text knowledge mining mechanism. Taking inspiration from CLIP, Song et al.~\cite{Song2022VisionLanguagePF} formulated three pretraining tasks for fine-grained cross-modality interaction, designed to align unimodal embeddings and learn enhanced representations of the backbone. Xue et al.~\cite{Xue2022LanguageMA} proposed a weakly supervised pretraining method, which simultaneously learns and aligns visual and partial text instance information, with the aim of producing effective visual text representations. 

    \subsection{Comparison to the Conference Version} 
    This paper is a substantial extension of our prior publication~\cite{Yu2023TurningAC}. Building upon this foundation, our current study incorporates three major improvements that contribute to the advancement of the field of scene text detection and scene text spotting.
    \begin{enumerate}
    
    \item We introduced FastTCM-CR50, an innovative text spotting backbone that overcomes the limitations of our conference version which is solely tested on scene text detection. It incorporates a meta query and Bimodal Similarity Matching (BSM), eliminating the need for text encoder in the inference process, leading to a remarkable speedup. Moreover, it dynamically augments text embeddings with visual modalities, enhancing the overall performance. Specifically, it brings about substantial improvements in inference speed (by 48.5\%) while enhancing performance.
    
    \item Extensive experiments were conducted to evaluate the performance of TCM and FastTCM in different settings. We explored their utility in boosting the efficacy of existing text detectors and spotters, their competence in few-shot learning, and their domain adaptation capabilities. Our thorough ablation studies offered insights into the contributions of our method in harnessing pretrained CLIP knowledge to elevate the performance of text detectors and spotters.

    \item Our method exhibited impressive adaptability across diverse tasks. The proposed FastTCM-CR50 showed their efficacy in scene text spotting and complex tasks like oriented, dense, and small object detection in aerial imagery. 
    
\end{enumerate}

\section{Methodology}
    \label{sec:method}
\begin{figure}[htbp]
        \centering
        \includegraphics[width=0.46\textwidth]{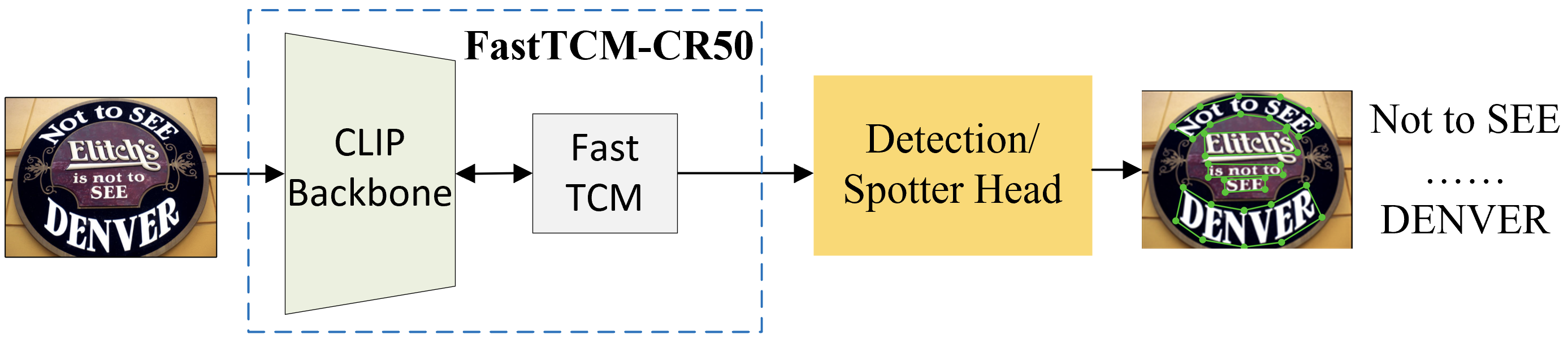}
        \caption{The overall framework of our approach.}
        \label{fig:method_overall}
    \end{figure}

     An overview of our approach is shown in Fig.~\ref{fig:method_overall}. In essence, we repurpose the CLIP model to serve as the backbone, utilizing the FastTCM as a bridge between the CLIP backbone and the detection/spotter heads. 

\subsection{Prerequisite: CLIP Model} 
    The CLIP model~\cite{Radford2021LearningTV} has demonstrated substantial potential in the realm of learning transferable knowledge and open-set visual concepts, given its capacity to analyze 400 million unannotated image-text pairs during its pretraining phase. Prior research~\cite{goh2021multimodal} reveals that CLIP's individual neurons are adept at capturing concepts in literal, symbolic, and conceptual manners, which serves as an innately text-friendly model, capable of effectively mapping the space between image and text~\cite{Petroni2019LanguageMA}. During its training phase, CLIP learns a joint embedding space for two modalities through a contrastive loss. Given a batch of image-text pairs, the model maximizes the cosine similarity with matching text and minimizes the similarity with all other unmatched text for each image. The same process applies to each piece of text, which has allowed CLIP to be utilized for zero-shot image recognition~\cite{Zhou2022ConditionalPL}. However, leveraging the valuable insights generated by such a model presents two prerequisites. First, an effective method is required to access the prior knowledge stored within the CLIP model. Second, while the original model is designed to measure the similarity between a complete image and a single word or sentence, scene text detection and spotting usually involve numerous text instances per image, all of which need to be equivalently recalled.

\subsection{FastTCM}
    FastTCM, designed to enhance the CLIP model, serves as a robust foundation for boosting existing scene text detectors and spotters. It achieves this by extracting both image and text embeddings from CLIP's image and text encoders, respectively. The first step in the process is designing a cross-modal interaction mechanism. We do this via visual prompt learning which restores the locality feature from CLIP's image encoder. The enhanced locality feature allows for capturing fine-grained data to effectively respond to a more general text region, setting the stage for subsequent matches between text instances and language. Next, to better channel pre-trained knowledge, we build a language prompt generator. This generator produces a contextual cue for each image. For the efficient extraction of interactions between the image and text encoder, all while enabling faster inferences, we use a method called Bimodal Similarity matching. This method allows for the offline computation of inferences using the CLIP text encoder, along with the dynamic generation of language prompts that are based on the conditions of the image. Finally, an instance-language matching technique is employed to align the image and text embeddings. This encourages the image encoder to meticulously refine text regions from the cross-modal visual-language priors.  
    
    \begin{figure}[htbp]
        \centering
        \includegraphics[width=0.5 \textwidth]{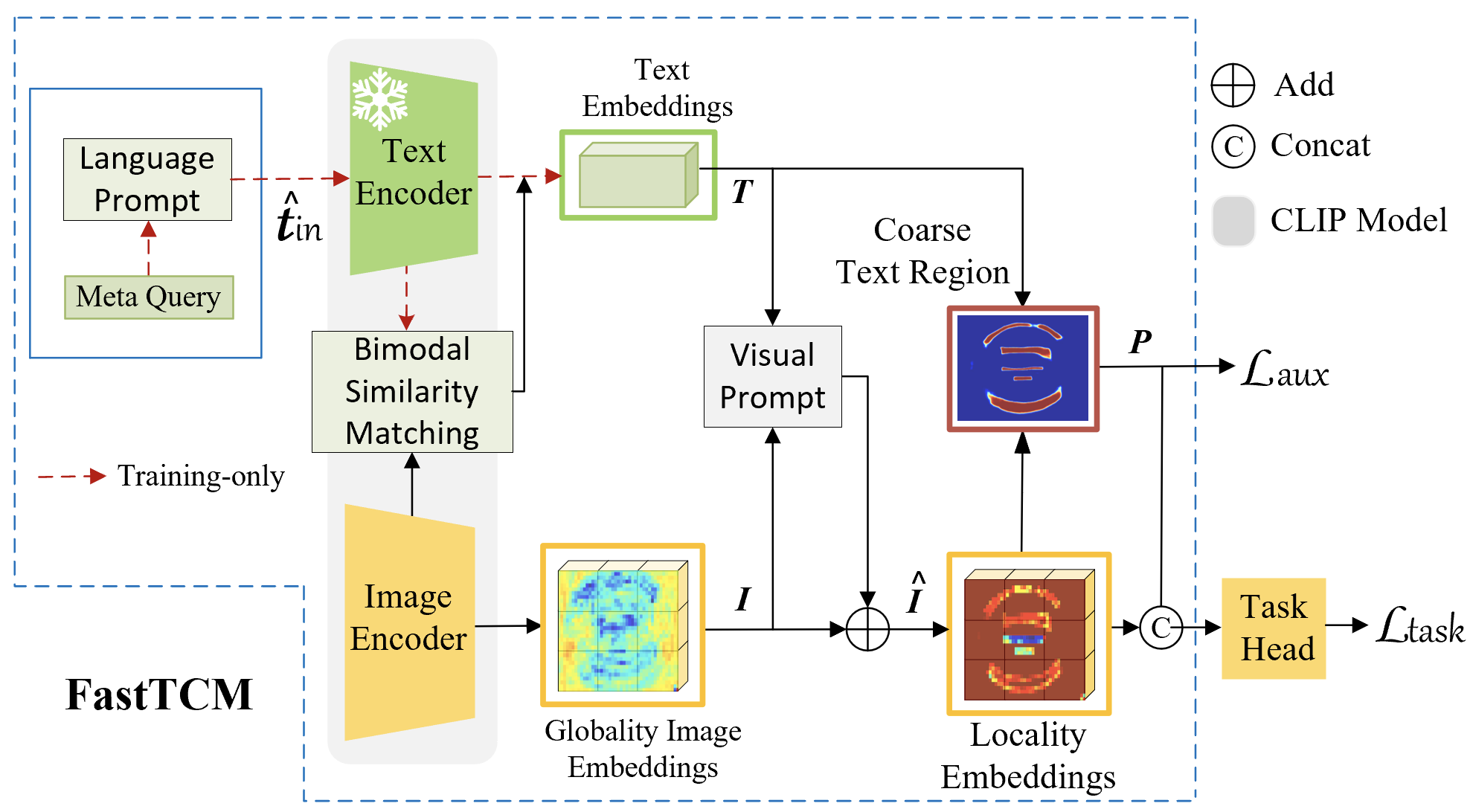}
        \caption{The details of the FastTCM. The image encoder and text encoder are directly from the CLIP model. The red dashed arrows represent training-only operators, with the corresponding upstream calculation procedure performed offline during the inference stage.}
        \label{fig:method_plug_tcm_spotter}
    \end{figure}
    
    \begin{figure}[htbp]
        \centering
        \includegraphics[width=0.4\textwidth]{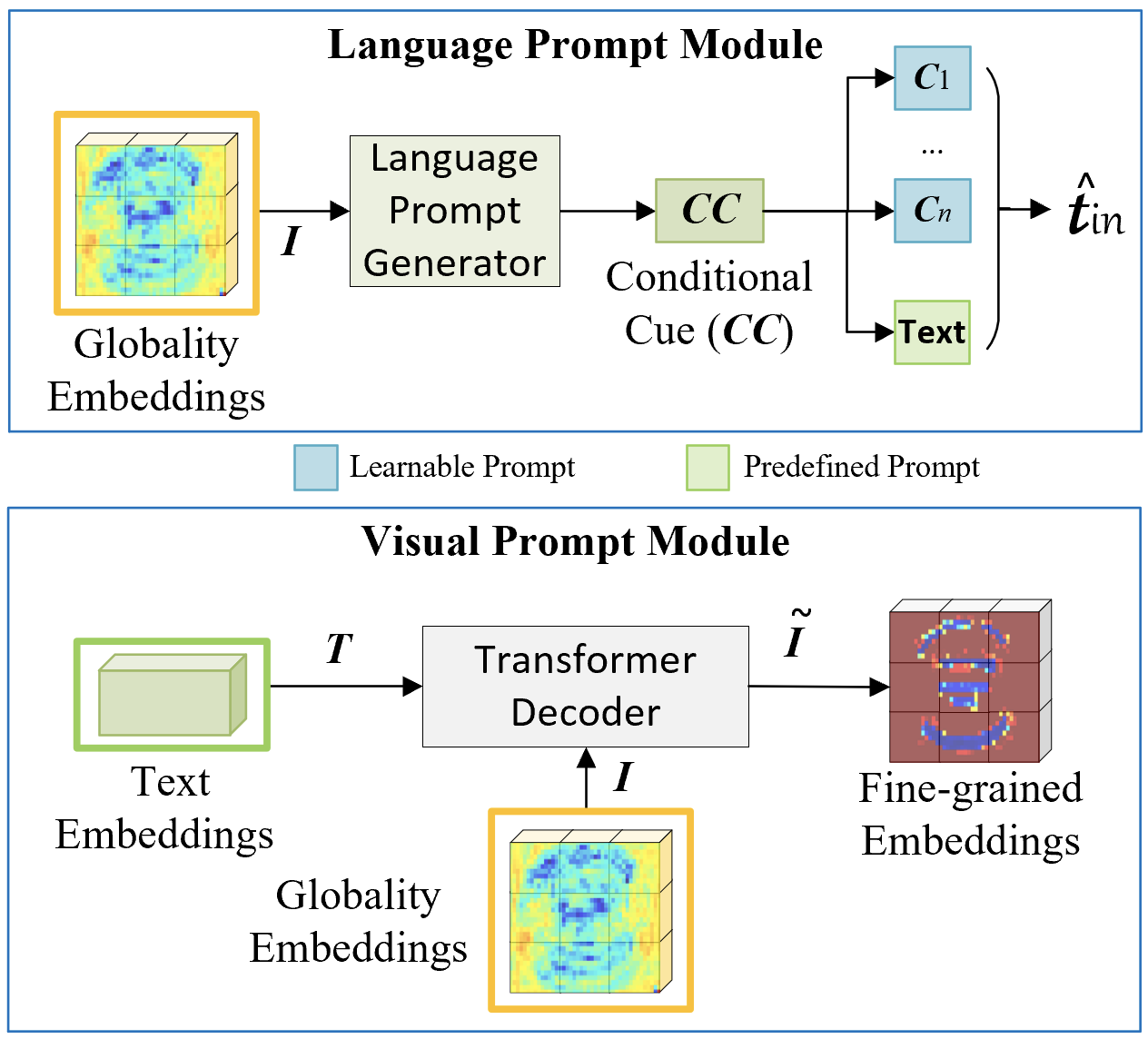}
        \caption{Illustration of the language prompt module (top) and visual prompt module (bottom).}
        \label{fig:method_plug_tcm_spotter_vis_lan_prompt}
    \end{figure}

    \subsubsection{Image Encoder} 
    We use the pretrained ResNet50~\cite{He2016DeepRL} of CLIP as the image encoder, which produces an embedding vector for every input pixel.
    Given the input image $ \bm{I}' \in \mathbb{R}^{ H \times W \times 3} $, image encoder outputs image embedding $\bm{I} \in \mathbb{R}^{\tilde{H} \times \tilde{W} \times C}$, where $\tilde{H} = \frac{H}{s}$, $\tilde{W} = \frac{W}{s}$, and $C$ is the image embedding  dimension ($C$ is set to 1024) and $s$ is the downsampling ratio (s is empirically set to 32), %
    which can be expressed as:
    \begin{equation} \label{eq:image_encoder}
    \bm{I} =  \operatorname{ImageEncoder}(\bm{I}')\,.
    \end{equation}

    \subsubsection{Text Encoder} The text encoder takes input a number of of $K$ classes prompt and embeds it into a continuous vector space $\mathbb{R}^C$, producing text embeddings $ \bm{T} = \{\bm{t}_1,\ldots,\bm{t}_K\} \in \mathbb{R}^{K \times C}$ as outputs of the text encoder, where $ \bm{t}_i \in \mathbb{R}^C$. Specifically, we leverage the frozen pretrained text encoder of CLIP throughout as the text encoder can provide language knowledge prior to text detection and spotting. $K$ is set to 1 because there is only one text class in text detection task. Different from the original model that uses templates like ``a photo of a [CLS].'', we predefine discrete language prompt as ``\emph{Text}''. Then, a part of the text encoder input $\bm{t}_{in}'$ is defined as follows:
    \begin{equation} \label{eq:image_text_encoder_input}
    \bm{t}_{in}' =  \operatorname{WordEmbedding}( \rm Text) \in \mathbb{R}^{D}\,,
    \end{equation}
    where $\operatorname{WordEmbedding}(\cdot)$ denotes word embedding for predefined prompt ``Text'' class. $D$ is the word embedding dimension and is set to 512.
    
    Inspired by CoOp~\cite{Zhou2021LearningTP, Zhou2022ConditionalPL}, we also add learnable prompt $\{\bm{c}_1,\ldots,\bm{c}_n\}$ to learn robust transferability of text embedding for facilitating zero-shot transfer of CLIP model,  where $n$ is the number of learnable prompt, which is set to 4 by default, and $\bm{c}_i \in \mathcal{R}^D$. Thus, the input $\bm{t}_{in}$ of the text encoder is as follows: 
    \begin{equation} \label{eq:t_in}
    \bm{t}_{in} = [\bm{c}_1,\ldots,\bm{c}_n, \bm{t}_{in}'] \in \mathbb{R}^{(n+1) \times D}\,.
    \end{equation}
     The text encoder takes $\bm{t}_{in}$ as input and generates text embedding $ \bm{T} = \{\bm{t}_1\} \in \mathbb{R}^{C}$, and $ \bm{T}$ is donated by $\bm{t}_{out} \in \mathcal{R}^C$ for simplification:
     \begin{equation} \label{eq:text_encoder}
     \bm{t}_{out} = \operatorname{TextEncoder}( \bm{t}_{in}) \in \mathbb{R}^{C}\,.
    \end{equation}

\subsubsection{Language Prompt Generator}
    Although the predefined prompt and learnable prompt are effective for steering the CLIP model, it may suffer from limited few-shot or generalization ability to open-ended scenarios where the testing text instance is out-of-distribution from the training images. 
    To this end, we present a language prompt generator to generate a feature vector, termed as conditional cue ($\bm{cc}$), as depicted in Fig.~\ref{fig:method_plug_tcm_spotter_vis_lan_prompt}. For each image,
    the $\bm{cc}$ is then combined with the input of the text encoder $\bm{t}_{in}$, formulated as follows:
    \begin{equation} \label{eq:text_gen}
    \hat{\bm{t}}_{in} = \bm{cc} + \bm{t}_{in} \in \mathbb{R}^{(n+1) \times D}\,,
    \end{equation}
    where $\hat{\bm{t}}_{in}$ is the new prompt input of the text encoder conditioned on the input image, and we replace $\bm{t}_{in}$ with $\hat{\bm{t}}_{in}$ in Eq.~\ref{eq:text_encoder}.

As depicted in Fig.~\ref{fig:method_plug_tcm_spotter},
we introduce meta query to generate an implicit conditional cue ($\bm{{cc}}$) followed by a two-layer feed-forward network, enabling the decoupling of the text encoder from the inference process. In addition, we design a bimodal similarity matching (BSM) module to act as a gate, which controls the amount of visual modal information that should compensate for text modal embeddings. This dynamic enrichment of text embeddings with visual information is helpful to the overall performance of the model. %

\noindent\textbf{Meta Query.} Specifically, FastTCM first incorporates a meta query, denoted as $\bm{MQ}$, which is initialized with learnable parameters representing the shape of $\mathbb{R^{C}}$. The meta query serves as an implicit image condition to guide the generation of subsequent language prompt, steering the pretrained knowledge from the text encoder. This operation is motivated by DETR~\cite{Carion2020EndtoEndOD}, which utilizes a Transformer Encoder, and Decoder that looks for a specific number of object queries (potential object detections). This substitution allows us to generate an implicit conditional cue $\bm{{cc}}$, and is formulated as follows:
\begin{equation} \label{eq:new_language_prompt_generator}
\bm{{cc}} = \operatorname{LN}(\sigma(\operatorname{LN}(\bm{MQ})\bm{W}_1+\bm{b}_1)) \bm{W}_2+\bm{b}_2 \in \mathbb{R}^{D}\,,
\end{equation}
where $\bm{{cc}}$ represents the generated implicit conditional cue, which is utilized in subsequent steps. $\bm{W}_1 \in \mathbb{R}^{ C \times C}$,  $\bm{W}_2 \in \mathbb{R}^{ C \times D}$, $\bm{b}_1 \in \mathbb{R}^{ C }$, $\bm{b}_2 \in \mathbb{R}^{ D }$, and we broadcast $\bm{cc}$ with $\bm{t}_{in}$ to get $\hat{\bm{t}}_{in}$ in Eq.~\ref{eq:text_gen}. It is important to note that once training is completed, the meta query remains unchanged. This allows the CLIP text encoder to perform offline participant calculation during inference, resulting in reduced inference time and making FastTCM more suitable for practical real-world applications. %

\noindent\textbf{Bimodal Similarity Matching.} Given the output of the text encoder $\bm{t}_{out}$ and the global image-level feature $\bm{\bar{I}}$, we first calculate the cosine similarity between text embeddings and globality image, as defined by the following equation:
\begin{equation} \label{eq:calculate_text_img_similarity}
sim = \frac{\bar{\bm{I}} \cdot \bm{t}_{out}}{|\bar{\bm{I}}| |\bm{t}_{out}|} \,,
\end{equation}
where $sim$ serves as the relevance threshold for an output gate that controls the amount of visual modal information used to compensate for text modal embeddings. Next, using the relevance threshold $sim$, we apply a weighted sum between $\hat{\bm{t}}_{out}$ and $\bar{\bm{I}}$ as follows:
\begin{equation} \label{eq:output_gate}
\hat{\bm{t}}_{out} = sim \cdot \bar{\bm{I}} + \bm{t}_{out} \,,
\end{equation}
where $\hat{\bm{t}}_{out}$ is the new output of the text encoder, which is dynamically post-conditioned on the implicit image features. We use $\hat{\bm{t}}_{out}$ to replace $\bm{t}_{out}$ in subsequent processes, including visual prompt generator (Eq.~\ref{eq:vis_prompt_gen}) and instance-language matching (Eq.~\ref{eq:updated_pixel_text_matching}).

\subsubsection{Visual Prompt Generator}
We design a visual prompt generator to adaptively propagate fine-grained semantic information from textual features to visual features, as presented in Fig.~\ref{fig:method_plug_tcm_spotter_vis_lan_prompt}. 
Formally, we use the cross-attention mechanism in Transformer~\cite{Vaswani2017AttentionIA} to model the interactions between image embedding ($\bm{Q}$) and text embedding ($\bm{K}$, $\bm{V}$). The visual prompt $\tilde{\bm{I}}$ is then learned for transferring the information prior from image-level to text instance-level, which is defined as:
 \begin{equation} \label{eq:vis_prompt_gen}
 \tilde{\bm{I}} = \operatorname{TDec}( Q= \bm{I}, K= \bm{t}_{out}, V= \bm{t}_{out} ) \in \mathbb{R}^{ \tilde{H} \times \tilde{W} \times C}\,,
\end{equation}
where TDec denotes the Transformer Decoder. In practice, it consists of 6 bidirectional transformer decoder layers with 4 heads for adequately interacting between image embeddings and text embeddings; transformer width is 256, and the feed-forward hidden dimension is set to 1024.

Based on the conditional visual prompt, the original image embedding $\bm{I}$ is equipped with  
$\tilde{\bm{I}}$ to produce the prompted text-aware locality embeddings $\hat{\bm{I}}$ used for instance-language matching~(Eq.~\ref{eq:updated_pixel_text_matching}) and downstream detection and spotting head:
\begin{equation} \label{eq:vis_gen}
\hat{\bm{I}} = \bm{I} + \tilde{\bm{I}} \,.
\end{equation}

\subsubsection{Instance-language Matching}
Given the output of the text encoder and image encoder, we perform text instance-language matching alignment on text-aware locality image embedding $\hat{\bm{I}}$ and text embedding $\bm{t}_{out}$ by the dot product followed by sigmoid activation to get binary score map. The mixture of the generated conditional fine-grained embedding $\tilde{\bm{I}}$ and visual embedding $\bm{I}$ can allow text instances existing in visual features to be better matched with pretrained language knowledge in collaboration. The matching mechanism is formulated as follows:
\begin{equation} \label{eq:updated_pixel_text_matching}
\bm{P} = \operatorname{sigmoid}(  \hat{\bm{I}}\bm{t}_{out}^T / \tau ) \in \mathbb{R}^{\tilde{H} \times \tilde{W} \times 1} \,,
\end{equation}
where $\bm{t}_{out}$ is text embedding because of only one text class in text detection scenarios, and $\tau$ is the temperature coefficient which is empirically set to 0.07 by default. $\bm{P}$ is the binary text segmentation map. The segmentation maps are supervised using the ground-truths as an auxiliary loss and concatenated by the prompted embedding $\hat{\bm{I}}$ for downstream text detection and spotting head to explicitly incorporate language priors for detection. During training, we minimize a binary cross-entropy loss between the segmentation map $\bm{P}$ and ground-truth, which is defined as follows:
\begin{equation} \label{eq:l_aux}
\mathcal{L}_{aux} = {\sum_i^{\tilde{H}}}\sum_j^{\tilde{W}} y_{ij}\log(P_{ij}) + (1-y_{ij})\log(1-P_{ij})\,,
\end{equation}
where $y_{ij}$ and $P_{ij}$ are the label and predicted probability of pixel $(i,j)$ belonging to the text instances, respectively.

\subsection{Optimization}
The loss function $\mathcal{L}_{total}$ is the sum of task loss $\mathcal{L}_{task}$ and auxiliary loss $\mathcal{L}_{aux}$, formulated as follows:
\begin{equation} \label{eq:total_loss}
\mathcal{L}_{total} = \mathcal{L}_{task} + \lambda \mathcal{L}_{aux} \,,
\end{equation}
where $\lambda$ is a trade-off hyper-parameters and set to 1 in this paper. $\mathcal{L}_{task}$ depends on downstream text detection methods including segmentation and regression categories, or text spotting methods. In the inference period, we use the output of the corresponding task head as the final result. In practice, we integrate the proposed method into both text detectors and text spotters to validate the effectiveness of our methods.

\section{Experiments}
\label{sec:experiments}
    We conduct extensive experiments to validate FastTCM. Our first set of experiments examines how FastTCM-CR50 backbone can be incorporated into existing text detectors and spotters to achieve consistent performance improvements. Next, we demonstrate the few-shot training capability and generalization ability by incorporating the FastTCM method. In the third set of experiments, we compare our method with previous pretraining methods tailored for text detection and spotting. Then, we provide thorough experiments to evaluate the sensitivity w.r.t. the proposed designs. Finally, we also conducted experiments on challenging oriented aerial object detection datasets to demonstrate the effectiveness of our method.       
    
    \subsection{Datasets} Our experiments are conducted on a number of commonly known scene text detection and spotting benchmarks including ICDAR2013 (IC13)~\cite{Karatzas2013ICDAR2R}, ICDAR2015 (IC15)~\cite{karatzas2015icdar}, MSRA-TD500 (TD)~\cite{yao2012detecting}, CTW1500 (CTW)~\cite{liu2019curved}, Total-Text (TT)~\cite{ch2019total}, ArT~\cite{chng2019icdar2019}, MLT17~\cite{Nayef2017ICDAR2017RR}, MLT19~\cite{nayef2019icdar2019}, SynthText~\cite{gupta2016synthetic}, CurvedSynthText-150k~\cite{Liu2020ABCNetRS}, and TextOCR~\cite{Singh2021TextOCRTL}. More details of the datasets refer to appendix.

\subsection{Implementation Details}
    In our text detection task experiments, we test the efficacy of prominent detection methodologies including DBNet 
    (DB)~\cite{Liao2020RealtimeST}, PAN~\cite{Wang2019EfficientAA}, and FCENet (FCE)~\cite{zhu2021fourier}. 
    The detection head from DBNet, PAN, and FCENet are utilized to yield the final results. To test the model's few-shot learning, we train on the benchmark using varying proportions of training data, and evaluate it against the corresponding test data. The generalization capability is tested by training it on respective source datasets and subsequently evaluating it on a target dataset with a markedly different distribution. The generalization ablity  of the FastTCM-CR50 is assessed through two different forms of adaptation: synthtext-to-real and real-to-real. A series of ablation studies are undertaken, focusing on the predefined prompt, the learnable prompt, the language prompt generator, the visual prompt generator, the BSM module, and various settings.

    For end-to-end text spotting tasks, we carry out experiments with recent methods such as Mask TextSpotter v3 (MTSv3)~\cite{Liao2020MaskTV}, ABCNet~\cite{Liu2020ABCNetRS}, ABINet++~\cite{Fang2022ABINetAB}, TESTR~\cite{Zhang2022TextST}, and DeepSolo~\cite{Ye2022DeepSoloLT}. These selected methods encompass both RoI-based and RoI-free text spotters. To ensure consistency with these text spotting methods, we use the same training approach, respecting the training data and hyper-parameters specific to each method.

\begin{table}[htbp]
  \centering
          \caption{
    Text detection results of cooperating with existing detectors on IC15, TD, and CTW. $^\dagger$ indicates the results from~\cite{Zhan2019GADANGD}.  ``BB'' denotes backbone where R50, CR50, TCM-CR50, FastTCM-CR50 represent the original ResNet50 backbones,  the pretrained CLIP ResNet50 backbone, the TCM-CR50 backbone, and ours FastTCM-CR50  backbone respectively. F (\%) represents F-measure. $\Delta $ means the improvement of performance between the cooperated method and the original method. FPS are reported using a single V100. %
}
  \begin{subtable}{1\linewidth}  %
    \centering
    \label{subtable1}
    \setlength\tabcolsep{1.8pt}

\input{table/text_det_ic15_td_ctw}

  \end{subtable}

  \label{tab:det_and_det_only_res}
\end{table}

\subsection{Cooperation with Existing Detector Methods}
    We assessed the impact of substituting the original backbones (ResNet50) of FCENet, PAN, and DBNet with the pretrained image encoder ResNet50 from CLIP (CR50). Yet, as evidenced in Tab.~\ref{tab:det_and_det_only_res}, merely leveraging the pretrained visual-language knowledge of the CLIP model (CR50) is inadequate for boosting scene text detection performance. This suggests the necessity of employing an appropriate method to harness the potential of the CLIP model.
    Subsequently, we evaluated the performance of FastTCM-CR50 with these two backbones. As illustrated in Tab.~\ref{tab:det_and_det_only_res}, FastTCM-CR50 can be effectively employed to augment current scene text detectors, yielding an average improvement of 1.7\% compared to the respective baseline methods. Furthermore, it is demonstrated that the FastTCM-CR50 backbone surpasses the TCM-CR50 backbone in terms of F-measure, contributing to an average performance enhancement of 0.2\% for DBNet, FCENet, and PAN on the IC15, TD, and CTW datasets, with an average speed improvement of 50.04\%.

\begin{figure}[ht]
    \centering
    \includegraphics[width=0.48\textwidth]{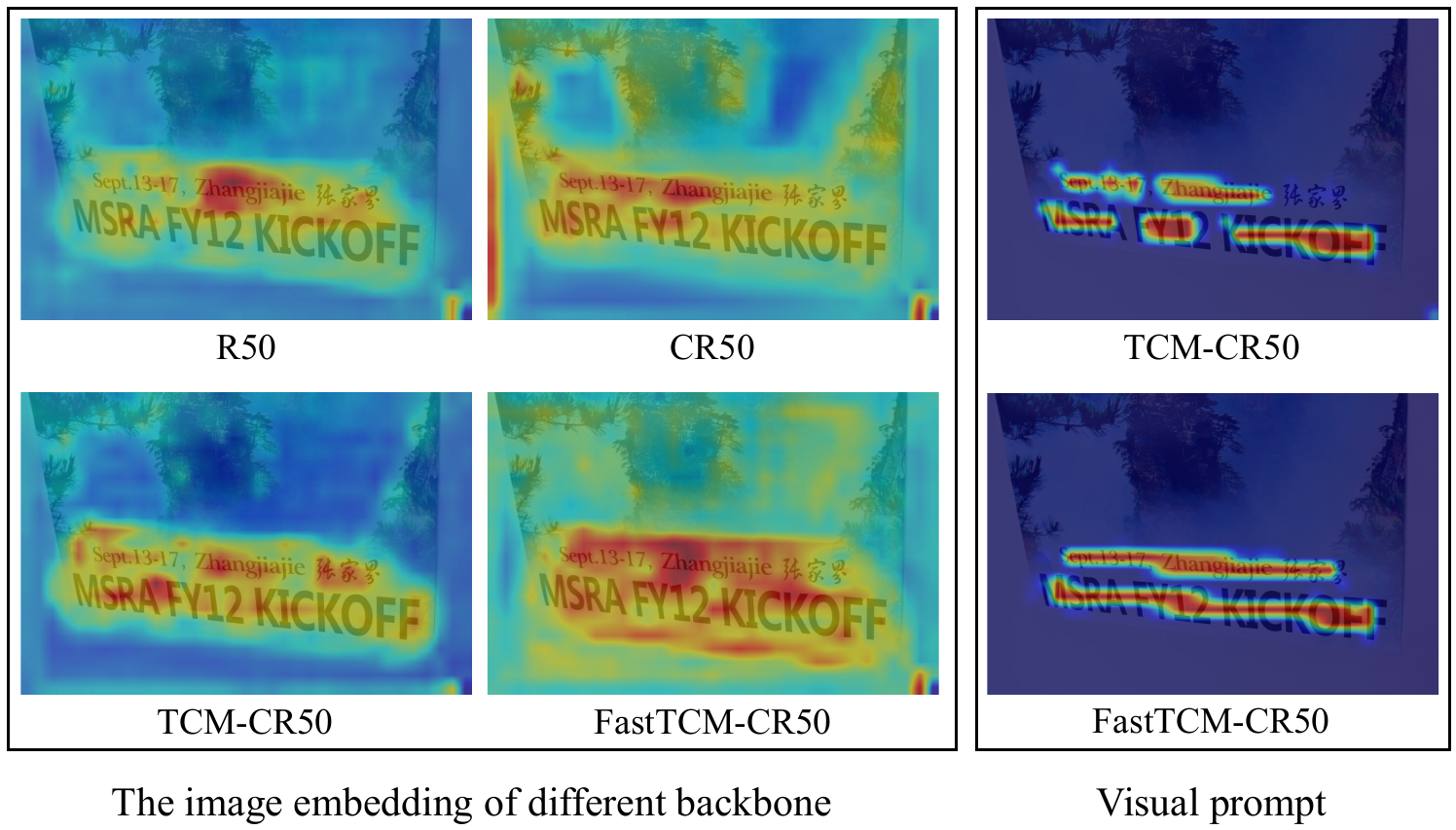}
    \caption{Visualization results of our method. The left is the image embedding $\bm{I}$ of different backbone including R50, CR50, TCM-CR50, and FastTCM-CR50, and the right is the generated visual prompt $\tilde{\bm{I}}$. Our method FastTCM-CR50 can accurately identify text regions. Best view in screen.}
    \label{fig:vp_results}
    \end{figure}

    We visualize our method in Fig.~\ref{fig:vp_results}. It shows that the fine-grained features $\tilde{\bm{I}}$ containing text information is recovered from the global image embedding $\bm{I}$, demonstrating that FastTCM can identify text regions and provide these prior cues for downstream text perception related tasks.

\subsection{Cooperation with Existing Spotter Methods}
\noindent \textbf{Detection-only Results.}
    As demonstrated in Tab.~\ref{tab:text_spot_and_det_only_tt_ic15_ctw_td}, we noted consistent enhancements in F-measure on text spotting benchmarks when TCM-CR50 was combined with five distinct text spotting methods. Particularly, TCM-CR50 outperformed the baseline methods such as MTSv3, ABINet++, ABCNet, DeepSolo, and TESTR with an R50 backbone, with performance boosts ranging from +0.2\% to +1.8\% in terms of F-measure on the TT dataset. Consistent improvements were also witnessed on IC15 and CTW datasets, underscoring TCM-CR50's suitability for text-spotting methods. Furthermore, when FastTCM-CR50 was integrated with MTSv3, ABINet++, ABCNet, DeepSolo, and TESTR, we observed an average performance enhancement of 0.2\% compared to TCM-CR50 based methods, accompanied by similar speed improvements, indicating FastTCM-CR50's superior efficacy. Additionally, the inclusion of an extra large-scale dataset, TextOCR, resulted in further performance gains, such as a 0.9\% improvement on the TT dataset using TESTR.

\noindent \textbf{End-to-end Spotting Results.}
    In Tab.~\ref{tab:text_spot_and_det_only_tt_ic15_ctw_td}, we present the end-to-end spotting performance of our method combined with existing scene text spotters. TCM-CR50 demonstrates favorable performance when integrated with various cooperative methods. Specifically, under the end-to-end setting with the strong lexicon on dataset IC15, TCM-CR50 outperforms the original MTSv3, ABINet++, ABCNet, DeepSolo, and TESTR by +0.8\%, +0.3\%, +2.3\%, +0.1\%, and +0.4\%, respectively, in terms of the `S' metric. Similar consistent improvements are also observed for datasets TT and CTW, indicating that TCM-CR50 effectively enhances the performance of both existing scene text detectors and spotters. Furthermore, when replacing TCM-CR50 with FastTCM-CF50, we observe a further improvement in performance, with an average gain of 1.5\% compared to baseline methods and an average gain of 0.56\% compared to TCM-CR50. Additionally, the inference speed of FastTCM-CR50 is increased by approximately 46.4\%. These results highlight the superiority of FastTCM-CR50 and its potential for efficient and accurate text spotting tasks. Besides, when using additional large-scale TextOCR as training data, our model can achieve further improvement, suggesting the compatibility of our method with large-scale datasets.

	\begin{table*}[ht]
		\centering
		\renewcommand\arraystretch{1}
		\caption{End-to-end text spotting results of cooperating with existing spotter methods on Total-Text, ICDAR2015, and CTW1500. ``Ext.'' short for extra data TextOCR. ``S'', ``W'', and ``G'' donate using strong, weak, and generic lexicons, respectively.
		F (\%) represents F-measure. $\Delta $ means the improvement of performance between the cooperated method and the original method. 
		}
		\setlength{\tabcolsep}{2.0pt}

\input{table/text_spot_and_det_only_tt_ic15_ctw_td}

		\label{tab:text_spot_and_det_only_tt_ic15_ctw_td}
	\end{table*}

\subsection{Few-shot Training Ability}
\begin{figure}[htbp]
    \centering
    \subcaptionbox{TD500}{\includegraphics[width=4.3cm,height=4.3cm]{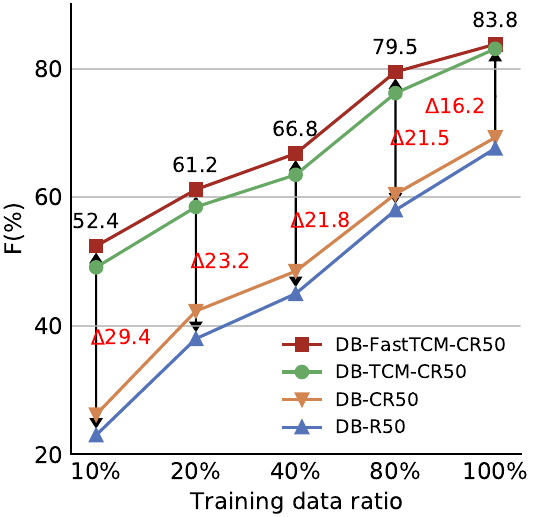}\label{fig:conditional_cue}}
    \subcaptionbox{ICDAR15}{\includegraphics[width=4.3cm,height=4.3cm]{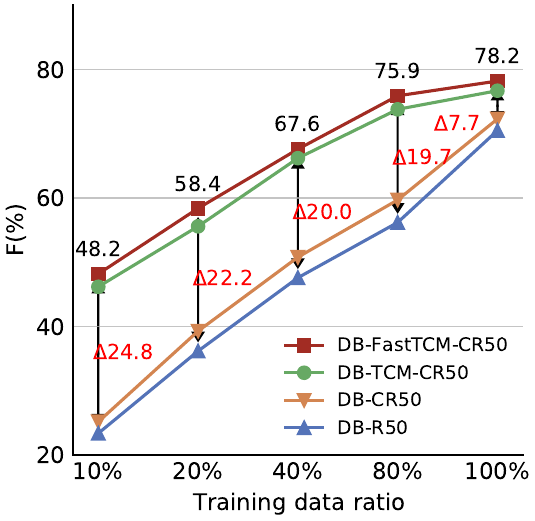}\label{fig:training_ratio_b}}
    \\
    \subcaptionbox{TotalText}{\includegraphics[width=4.3cm,height=4.3cm]{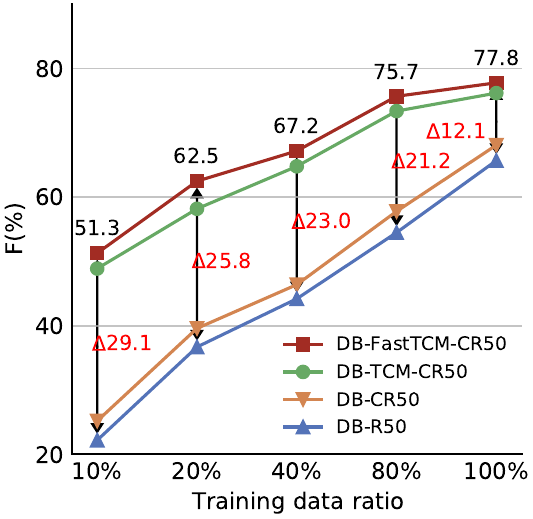}\label{fig:training_ratio_c}}
    \subcaptionbox{CTW1500}{\includegraphics[width=4.3cm,height=4.3cm]{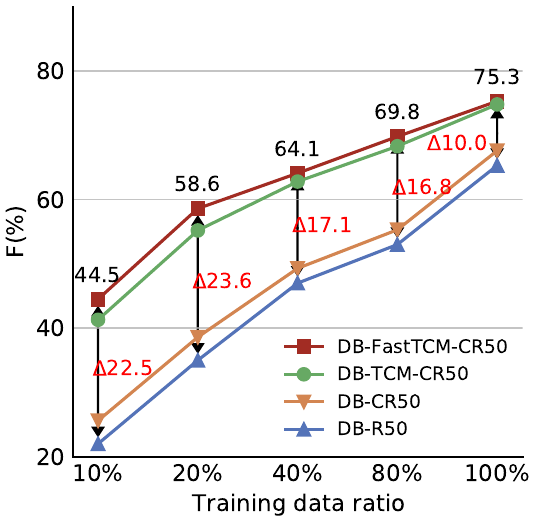}\label{fig:training_ratio_d}}
    \caption{Few-shot training ability of text detection task with varying training data ratio. ``F'' represents F-measure. 
    }
    \label{fig:few_show_ability}
\end{figure}

    \noindent \textbf{Results for Text Detection Task.} To verify the few-show training ability of our method on text detection tasks, we directly train our model on real datasets using various training data ratios without pretraining, and evaluate it on the corresponding 4 benchmarks. As shown in Fig.~\ref{fig:few_show_ability}, DB-FastTCM-CR50 shows robustness on limited data and outperforms the baseline methods DB in an average of 26.5\% in terms of 10\% training data ratio settings. Besides, DB-CR50 has limited improvements compared to our specific design FastTCM. The results show that the FastTCM can capture the inherent characteristic of text via leveraging the pretrained vision and language knowledge of the zero-shot trained CLIP model.

    \noindent \textbf{Few-shot Experiments for Text Spotting.} In addition, we performed few-shot experiments on text spotting tasks using ABCNet and TESTR on Total-Text, as illustrated in Tab.~\ref{tab:few_shot_on_spotting}. Considering that the recognizer module in text spotting methods often struggles to learn effectively with very limited data, we followed the text spotting pretraining step to obtain a suitable initialization for the corresponding text spotting methods. Subsequently, we applied different training ratios of the Total-Text dataset to evaluate the few-shot learning ability. The results demonstrate that both TCM-CR50 and FastTCM-CR50 exhibit advantages in few-shot learning for text spotting tasks compared to DB-R50 and DB-CR50. Moreover, using our method outperforms baseline methods by an average of 5.5\%. This demonstrates the effectiveness and superiority of FastTCM-CR50 over simply replacing other counterparts.

    \begin{table}[htbp]
	
        \setlength\tabcolsep{1pt}
		\centering
  		\caption{Few-shot training ability of text spotting task with varying training data ratio on Total-Text.
		 End-to-End spotting metric ``None'' (\%) is reported. %
		}
		\renewcommand\arraystretch{1}

		\setlength{\tabcolsep}{2.0mm}
		\resizebox{\linewidth}{!}{

  \newcommand{\tabincell}[2]{\begin{tabular}{@{}#1@{}}#2\end{tabular}}
\begin{tabular}{rcccccc}
    \toprule
	\multirow{1}[0]{*}{Method} & 
 	\multirow{1}[0]{*}{BB} & 
        \multirow{1}[0]{*}{\tabincell{c}{10\%}} & 
	\multicolumn{1}{c}{20\%} &
	\multicolumn{1}{c}{40\%} &
	\multicolumn{1}{c}{80\%} &
	\multicolumn{1}{c}{100\%} 
	 \\
    \midrule
        \multirow{4}[2]{*}{ABCNet}  &R50 & 43.5  & 51.7  & 55.2 & 62.3 & 64.2 \\
         &CR50 & 46.2 & 53.5 & 58.6 & 63.7 & 65.3  \\
	   &TCM-CR50 & 50.6 & 57.2 & 63.2 & 69.3 & 73.4 \\
 	 &FastTCM-CR50 & \textbf{52.3} & \textbf{60.1} & \textbf{66.2} & \textbf{71.6} & \textbf{74.1} \\ \midrule
	\multirow{4}[2]{*}{TESTR} &R50 & 58.4  & 62.2  & 68.5 & 69.3 & 73.2 \\
	   & CR50 &59.1  & 63.9 & 68.7 & 70.4 & 73.4 \\
         & TCM-CR50 & 60.3 & 66.8 & 68.9 & 72.8 & 73.6 \\
 	 &FastTCM-CR50 & \textbf{60.5} & \textbf{67.0} & \textbf{69.2} & \textbf{73.4} & \textbf{74.3} \\
    \bottomrule
\end{tabular}

		}

		\label{tab:few_shot_on_spotting}
	\end{table}

\subsection{Generalization Ability}
\label{sec:exp:gn}

\noindent\textbf{CLIP Backbone Generalization.} We conducted an experiment to investigate the generalization performance of DBNet by directly replacing the backbone of DBNet with CLIP backbone (CR50), as shown in Tab.~\ref{tab:synth_to_real}. It shows that the CLIP-R50 can indeed bring benefits for generalization. However, by integrating with FastTCM-CR50 backbone, the performance can be significantly improved. It suggests that directly using the pretrained CLIP-R50 is not strong enough to improve the generalization performance of the existing text detector, which further indicates that synergistic interaction between the detector and the CLIP is important. Meanwhile, FastTCM-CR50 also consistently outperforms TCM-CR50.

\noindent\textbf{Synth-to-real and real-to-real Adaptation.} We conduct two types of experiments including synthtext-to-real adaptation and real-to-real adaptation on text detection tasks, as shown in Tab.~\ref{tab:synth_to_real} and Tab.~\ref{tab:real_to_real}, respectively. Real-to-real adaptation contains monolingual and multi-lingual scenarios.
From the tables, we can see that by integrating the FastTCM-CR50 into DBNet, we significantly improve the performance by an average of 12.4\% in terms of F-measure for four different settings including synthtext-to-real and real-to-real, which further demonstrates the effectiveness of our method for domain adaptation. Notably, FastTCM-CR50 also consistently demonstrates improvements by an average of 0.4\% compared to TCM-CR50 as well, further emphasizing the remarkable generalization ablity of our methods.

	\begin{table}[t]
		\centering
  		\caption{Synthtext-to-real adaptation. $^\dagger$ indicates the results from~\cite{Wu2020SynthetictoRealUD}. ST indicates SynthText. F-measure (\%) is reported.}
		\renewcommand\arraystretch{1}
		\setlength{\tabcolsep}{2.0mm}
  \resizebox{0.83\linewidth}{!}{
		\input{table/synth_to_real}

}
		\label{tab:synth_to_real}
	\end{table}

	\begin{table}[ht]
		\centering
  		\caption{Real-to-real adaptation. $^\dagger$ indicates that the results are from~\cite{Zhan2019GADANGD}. %
		Note that the proposed method outperforms other methods. F-measure (\%) is reported.
		}
		\renewcommand\arraystretch{1}
		\setlength{\tabcolsep}{0.5pt}
	    \resizebox{\linewidth}{!}{
		\input{table/real_to_real}

		}

		\label{tab:real_to_real}
	\end{table}

     \begin{table}[htbp]
	
        \setlength\tabcolsep{1pt}
		\centering
  		\caption{Real-to-real adaptation on scene text spotting methods
		End-to-End spotting metric `None' (\%) is reported. %
		}
		\renewcommand\arraystretch{1}

		\setlength{\tabcolsep}{0.7pt}
		\resizebox{\linewidth}{!}{
		
\newcommand{\tabincell}[2]{\begin{tabular}{@{}#1@{}}#2\end{tabular}}
\begin{tabular}{lccccc}
    \toprule
	\multirow{1}[0]{*}{Method} & 
 	\multirow{1}[0]{*}{BB} & 
	\multicolumn{1}{c}{ TT$\rightarrow$IC15 } &
	\multicolumn{1}{c}{TT$\rightarrow$CTW} &
	\multicolumn{1}{c}{IC15$\rightarrow$CTW} &
	\multicolumn{1}{c}{CTW$\rightarrow$IC15} 
	 \\

     \midrule
	\multirow{2}[1]{*}{ABCNet}    & R50 & 32.5  & 37.1 & 33.5 & 34.6 \\
	  &FastTCM-CR50 & \textbf{48.2}  &  \textbf{50.3 }&  \textbf{47.8 }&  \textbf{51.7} \\
     \midrule
	\multirow{2}[1]{*}{TESTR}   &R50 & 36.7  & 39.2 & 36.1 & 36.8 \\
	 &FastTCM-CR50  &  \textbf{53.1}  &  \textbf{49.2} &  \textbf{48.1 }&  \textbf{56.2} \\
 
    \bottomrule
\end{tabular}
  
  }

		\label{tab:domain_spotting}
	\end{table}

\noindent\textbf{Real-to-Real Adaptation on Scene Text Spotting.} Besides, we also conducted real-to-real adaptation experiments with existing spotting methods, as shown in Tab. \ref{tab:domain_spotting}. The results show that the FastTCM-CR50 has the capacity of improving the existing scene text spotting methods by an average of 14.8\%, further demonstrating the effective generalization ability.

\subsection{Comparison with Pretraining Methods}
\label{subset:pretrain}
The pretraining methods based on specifically designed pretext tasks have made effective progress in the field of text detection. In contrast to these efforts, FastTCM-CR50 can turn the CLIP model directly into a scene text detector without pretraining process. The comparison results are shown in Tab.~\ref{tab:comap_pretraining}, from which we can see that without pretext tasks for pretraining, DB+FastTCM-CR50 consistently outperforms previous methods including DB+STKM~\cite{Wan2021SelfattentionBT}, DB+VLPT~\cite{Song2022VisionLanguagePF}, and  DB+oCLIP~\cite{Xue2022LanguageMA}. Especially on IC15, our method outperforms the previous state-of-the-art pretraining method by a large margin, with 89.5\% versus 86.5\% in terms of the F-measure. 
Furthermore, we demonstrate the proposed backbone can also be further improved using such pretext tasks pretraining as in oCLIP, with an average of 0.11\% improvement in terms of the F-measure.

\begin{table}[tb]
\centering
\caption{Comparison with existing scene text pretraining techniques on DBNet (DB). $^\dagger$ indicates the results from~\cite{Song2022VisionLanguagePF}. ST and VLP denote SynthText pretraining and visual-language pretraining methods, respectively. * stand for the results from~\cite{Xue2022LanguageMA}. F-measure (\%) is reported. 
}
\normalsize
\setlength\tabcolsep{2.0pt}
{
\begin{tabularx}{1.0\linewidth}{l|lcccccc}

\toprule
                                                      & Method &BB   & IC15 & TT    & TD    & CTW   \\ \midrule
\multirow{5}{*}{\rotatebox{90}{Convention}}                  & SegLink~\cite{Shi2017DetectingOT} &VGG16                & -    & -     & 77.0    & -     \\
                                                      & PSENet-1s~\cite{Li2019ShapeRT} &R50           & 85.7 & 80.9  & -     & 82.2  \\
                                                      & LOMO~\cite{Zhang2019LookMT} &R50             & 87.2 & 81.6  & -     & 78.4  \\
                                                      & MOST~\cite{He2021MOSTAM}  &R50           & 88.2 & -     & 86.4  & -     \\
                                                      & Tang \etal\cite{Tang2022FewCB}  &R50         & 89.1 & -     & 88.1  & -     \\ \midrule
\multirow{4}{*}{\rotatebox{90}{VLP}} & DB+ST$^\dagger$  &R50                & 85.4 & 84.7  & 84.9  & -     \\
                                                      & DB+STKM$^\dagger$~\cite{Wan2021SelfattentionBT} & R50             & 86.1 & 85.5  & 85.9  & -     \\
                                                      & DB+VLPT$^\dagger$~\cite{Song2022VisionLanguagePF} &R50              & 86.5 & 86.3  & 88.5  & -     \\
                                                      & DB+oCLIP*~\cite{Xue2022LanguageMA} &R50             & 85.4   & 84.1     & -     & 82.0  \\ \midrule
                                                      & \multirow{1}[1]{*}{DB} &TCM-CR50        & {89.4} & 85.9 & {88.8} & {85.1} \\ 
                                                       & DB &FastTCM-CR50           & \textbf{89.5} & 86.1  & \textbf{88.9} & \textbf{85.2} \\ 
                                                          & DB+oCLIP*~\cite{Xue2022LanguageMA} &FastTCM-CR50     & \textbf{89.6} &86.2   & \textbf{88.94} & \textbf{85.4} \\ 
                                                       \bottomrule

\end{tabularx}}

\label{tab:comap_pretraining}
\end{table}

\subsection{Ablation Studies}

    \begin{table}[htbp]
        \centering
        		\caption{Ablation study of our proposed components on IC15, TD, TT and CTW. 
		``BSL'', ``PP'', ``LP'', ``LG'', ``VG'', ``Aux.'',  and ``BSM'' represent the baseline method DBNet, the predefined prompt, the learnable prompt, the language prompt generator, the visual prompt generator, auxiliary loss, and bimodal similarity matching, respectively. F (\%) represents F-measure. $\Delta $ represents the variance.
		}
        \normalsize
        \setlength\tabcolsep{0.8pt}

\input{table/ablation_ic15_td_tt_ctw}

		\label{tab:abla_ic15_td}
	\end{table}

\noindent\textbf{Ablation Study for the Predefined Prompt.} When using the predefined prompt, as illustrated in the second row of Tab.~\ref{tab:abla_ic15_td}, the performances are slightly improved on all four datasets (IC15, TD, TT, and CTW), with 0.05\%, 0.2\%, 0.04\%, and 0.1\% higher than the baseline method, respectively.

\noindent\textbf{Ablation Study for the Learnable Prompt.} Then, results combing the learnable prompt with the predefined prompt on four datasets are provided in the third row of Tab.~\ref{tab:abla_ic15_td}. We notice that a consistent improvement can be achieved by adding the learnable prompt.
We also show the influence of using different numbers of the learnable prompt in row 4 to row 6 of Tab.~\ref{tab:abla_ic15_td}. We observe that as the value of the number of the learnable prompt increases, the performance increases gradually on all datasets. Compared to the value 4, the value 32 obtains obvious improvements on CTW, TD, and TT. We conjecture that this is because the larger number of the learnable prompt can better steer the pretrained text encoder knowledge which is useful for text detection. In the following experiments, the default number of the learnable prompt is set to 4 for simplicity.

\noindent\textbf{Ablation Study for the Language Prompt Generator.} Besides, we evaluate the performance of the proposed language prompt generator shown in 7$_{th}$ row of Tab.~\ref{tab:abla_ic15_td}. 
With the help of the language prompt generator, we find that TCM achieves further improvements on all four datasets, especially on ICDAR2015, indicating that the conditional cue generated by the language prompt generator for each image can ensure better generalization over different types of datasets.

\noindent\textbf{Ablation Study for the Visual Prompt Generator.} Furthermore, combining the proposed visual prompt generator with the above other components, the improvement of F-measure is better than the baseline on all four datasets, with larger margins of 1.7\% and 2.0\% on IC15 and TD, respectively.
The reason for this obvious complementary phenomenon is that the visual prompt generator can propagate fine-grained visual semantic information from textual features to visual features. 
Besides, the prompted locality image embedding generated by the visual prompt generator can guide the model to obtain more accurate text instance-level visual representations, which boosts the ability of instance-language matching and generates a precise segmentation score map that is useful for downstream detection head.

\noindent\textbf{Ablation Study for the Bimodal Similarity Matching.} We further conducted a comparison of the results with and without bimodal similarity matching, as outlined in the last group of Tab.~\ref{tab:abla_ic15_td}. The results clearly demonstrate that the utilization of bimodal similarity matching leads to higher performance. This finding indicates that bimodal similarity matching plays a crucial role in training the model by dynamically enriching text embeddings with visual information, resulting in improved overall performance.

\noindent\textbf{Ablation Study for the Auxiliary Loss.}
We compare the results of with and without auxiliary loss, as shown in Tab.~\ref{tab:abla_ic15_td}. We observe that using auxiliary loss achieves higher performance. The results indicate auxiliary loss is beneficial to train the model via imposing constraints on instance-language matching score map. In addition, the improvement of the performance suggests that it might help the image encoder of pretrained CLIP to perceive locality text regions effectively.

\noindent\textbf{Ablation Study for the Key Component on Generalization Performance.} As presented in Tab.~\ref{tab:abla_da_tg_vg}, removing the meta query and BSM elements from FastTCM dramatically deteriorates the generalization performance, highlighting the importance and effectiveness of these components. Similarly, removing the VG and LG elements from FastTCM also results in a substantial drop in generalization performance, further validating their effectiveness. Finally, when we remove all of these components, the performance experiences an additional significant drop, indicating that each of these components contributes to the overall effectiveness and performance of FastTCM-CR50.

  	\begin{table}[htbp]
		\centering
  		\caption{
		Ablation study of the effect of meta query, BSM, LG, and VG on generalization performance. MQ is short for meta query. F-measure (\%) is reported.
		}
		\renewcommand\arraystretch{1}
		\setlength{\tabcolsep}{1.0pt}

\input{table/abla_da_tg_vg}

		\label{tab:abla_da_tg_vg}
	\end{table}

\noindent\textbf{Ablation Study for the Parameters Comparison.}
For a fair comparison, we have increased the parameters of DBNet by replacing the backbone with a larger ResNet and then conducting text detection experiments on TD dataset and a domain adaptation experiment on IC13 $\rightarrow$ IC15. Trainable parameters and FLOPs are calculated with an input size of 1280$\times$800. Results are shown in Tab.~\ref{tab:appendix_fair_db}.
The results show that DBNet with FastTCM-CR50 has better performance than DBNet with less model size and computation overhead compared to DBNet with R152 backbone, demonstrating its effectiveness.

	\begin{table}[ht]
		\centering
  		\caption{Ablation study of the trainable parameters comparison with DBNet on TD dataset and IC13 $\rightarrow$  IC15. ``Num.`` refer to the number of transformer decoder layers of VG. F-measure (\%) is reported.
		}
		\renewcommand\arraystretch{1}

		\setlength{\tabcolsep}{1.3pt}
		
\begin{tabular}{llcccccc}
\toprule
Method    & BB &Num. & Params & FLOPs & TD & IC13 $\rightarrow$  IC15 & FPS\\ \midrule
\multirow{4}[1]{*}{DB}    & R50  &-     &    26 (M)  &   98 (G)  & 84.9   &63.9 &14.5 \\
     & R101   &-  &    46 (M)  &   139 (G) & 85.9    &  64.3 & 11.7\\
     & R152   &-  &    62 (M)  &  180 (G)  & 87.3    &   64.7 & 8.4 \\ \midrule
\multirow{4}[1]{*}{DB} & TCM-CR50    &6   &    50 (M)  &  156 (G)  & {88.7} &71.9 &10\\ 
\cline{2-8}
 & FastTCM-CR50    & 1  &     30 (M)  &  107 (G)  &  88.1 &72.0 & 14.2 \\
  & FastTCM-CR50   & 3   &    34 (M)  &  117 (G)  &  88.5 & 72.2 & 13.8\\
 & FastTCM-CR50    &6  &    50 (M)  &  156 (G)  & \textbf{88.9}  &\textbf{72.4} & 13.3\\
\bottomrule
\end{tabular}

		\label{tab:appendix_fair_db}
	\end{table}

\noindent\textbf{Ablation Study for the number of transformer decoder layers of VG.}
The last group of Tab.~\ref{tab:appendix_fair_db} demonstrates the impact of varying the number of transformer decoder layers of VG on the performance. The results show that the performance remains robust across different numbers of decoder layers. This indicates that in practical applications, we have the flexibility to decrease the number of transformer decoder layers to achieve a better trade-off between model parameters and performance. By reducing the number of layers, we can potentially save computational resources and memory while maintaining satisfactory performance, making the model more efficient and practical for real-world applications.

\noindent\textbf{Ablation Study for the Different Predefined Language Prompt.} We conducted ablation study on the predefined language prompt with different strings using DBNet with FastTCM-CR50 in Tab.~\ref{tab:differ_predefined_prompt}. Results show that without predefined language prompt, the performance is harmed. In addition, it can be seen that there is little performance variation with different predefined language prompt. When the predefined language prompt becomes long and complex, the model performance drops a little. We deem that the CLIP is not good at handling complex instructions because it is pretrained on a dataset of 400 million image-text pairs that contain noise. As a result, this noise can affect the CLIP's ability to deal with long instructions.

\begin{table}[ht]
\centering
\caption{Ablation study of the different predefined language prompt with DBNet-FastTCM-CR50 on TD. F-measure (\%) is reported.}
\normalsize
\setlength\tabcolsep{2.3pt}
{
\begin{tabularx}{1.0\linewidth}{lc}
\toprule
Predefined language prompt       & TD   \\
\midrule
``Text''         &   \textbf{88.9}        \\
``A set of arbitrary-shape text instances'' & 88.7 \\ 
``The pixels of many arbitrary-shape text instances'' & 88.6 \\
w/o predefined language prompt & 87.4 \\
\bottomrule
\end{tabularx}}

\label{tab:differ_predefined_prompt}
\end{table}

\begin{table*}[tb]
\centering
\caption{Detection results of cooperating with existing rotated object detection methods on the DOTA-v1.0 testing set. R50, R101, and R152 denote ResNet-50, ResNet-101, and ResNet-152, respectively.
MS indicates that multi-scale testing is used. %
}
\normalsize
\setlength\tabcolsep{1.3pt}
{
\begin{tabularx}{1.0\linewidth}{c|lccccccccccccccccc|l}
\hline
 & Method & BB & MS & PL & BD & BR & GTF & SV & LV & SH & TC & BC & ST & SBF & RA & HA & SP & HC & mAP$_{50}$   \\
\hline
\multirow{14}*{\rotatebox{90}{Single-stage}} & PolarDet~\cite{PolarDet} & R101 & $\surd$ & 89.7 & 87.1 & 48.1 & 71.0 & 78.5 & 80.3 & 87.5 & 90.8 & 85.6 & 86.9 & 61.6 & 70.3 & 71.9 & 73.1 & 67.1 & 76.6 \\
& RDD~\cite{RDD} & R101 & $\surd$ & 89.1 & 83.9 & 52.5 & 73.1 & 77.8 & 79.0 & 87.1 & 90.6 & 86.7 & 87.1 & 64.1 & 70.3 & 77.0 & 75.8 & 72.2 & 77.8\\
& GWD~\cite{GWD} & R152 & $\surd$ & 89.1 & 84.3 & 55.3 & 77.5 & 77.0 & 70.3 & 84.0 & 89.8 & 84.5 & 86.1 & 73.5 & 67.8 & 72.6 & 75.8 & 74.2 & 77.4\\ \cline{2-20}
& \multirow{2}{*}{KLD~\cite{KLD}} & R50 & & 88.9 & 83.7 & 50.1 & 68.8 & 78.2 & 76.1 & 84.6 & 89.4 & 86.2 & 85.3 & 63.1 & 60.9 & 75.1 & 71.5 & 67.5 & 75.3\\
& ~ & R50 & $\surd$ & 88.9 & 85.2 & 53.6 & 81.2 & 78.2 & 77.0 & 84.6 & 89.5 & 86.8 & 86.4 & 71.7 & 68.1 & 76.0 & 72.2 & 75.4 & 78.3\\ \cline{2-20}

& \multirow{3}[1]{*}{RetinaNet-O~\cite{lin2017focal} }   & R50 & & 88.7 & 77.6 & 41.8 & 58.2 & 74.6 & 71.6 & 79.1 & 90.3 & 82.2 & 74.3 & 54.8 & 60.6  & 62.6 & 69.7 & 60.6 & 68.4 \\
&   & CR50 & & 87.9 & 74.7& 36.5 & 61.7 & 77.7 & 64.9 & 77.4 & 90.1 & 79.6 & 78.3 & 54.5 & 60.4 & 61.3 & 57.6 & 41.3 & 66.9 (\textbf{\redp{-1.5}})\\
&   & FastTCM-CR50 & & 87.5 & 77.9 & 41.0  & 66.2 & 70.8 & 72.5 & 78.3 & 89.9  & 81.6 & 83.8 & 55.7 & 60.3 & 62.6 & 71.0 & 58.1 & 70.5 (\textbf{\greenp{+2.1}}) \\ 

\cline{2-20} 

& \multirow{3}[1]{*}{{Rotated-FCOS}~\cite{tian2019fcos} }     & R50 & &89.2 & 72.0 & 48.0 & 61.6 & 79.3  & 73.5 & 85.8 & 90.9  & 81.1 & 84.3  & 59.6 & 62.7 & 62.1 & 69.9 & 49.3 & 71.3 \\
&   & CR50 & & 88.3 & 73.0 & 46.0 & 55.5 & 78.1 & 69.0 & 87.5 & 90.9 & 81.5 & 82.0& 57.7& 62.6 & 65.6 & 59.4 & 43.6 & 69.4 (\textbf{\redp{-1.9}})\\
&  & FastTCM-CR50 & &88.4 & 77.2 & 45.8 & 59.6 & 81.3 & 83.1 & 87.9 & 90.9 & 84.9 & 85.0 & 57.0 & 64.8  & 72.3 & 77.2 & 58.3 & 74.3 (\textbf{\greenp{+3.0}}) \\ 

\cline{2-20}
& \multirow{3}[1]{*}{{Rotated-ATSS}~\cite{zhang2020bridging}  }    & R50 & & 88.9 & 79.9 & 48.7 & 70.7 & 75.8  & 74.0 & 84.1 & 90.9 & 83.2 & 84.1 & 60.5 & 65.1 & 66.7 & 70.1 & 57.8 & 73.4 \\
&   & CR50 & & 89.1 & 75.2 & 45.9 & 66.8 & 78.4 & 74.7 & 87.2 & 90.8 & 83.1 & 84.8 & 53.5 & 65.1 & 69.6 & 64.6 & 52.8 & 72.1 (\textbf{\redp{-1.3}})\\
&  & FastTCM-CR50 & & 88.7 & 80.5 & 46.7 & 69.9 & 81.1 & 83.5 & 87.8 & 90.9 & 82.7 & 85.8 & 60.6 & 63.6 & 72.9 & 78.7 & 56.2 & 75.3 (\textbf{\greenp{+1.9}}) \\ 

\hline
\end{tabularx}}
\label{tab:dota}
\end{table*}

\noindent\textbf{Ablation Study for Different Amount of Data.}
To further explore whether the FastTCM can learn the additional knowledge which is hard to be obtained from increasing data, we have trained the model on large-scale public joint data including IC13, IC15, TD, CTW, TT, and MLT17, with a total of 13,784 images, and testing it on a NightTime-ArT data (326 images) carefully collected from ArT. The nighttime examples of ArT are provided in appendix. %
Results are shown in Tab.~\ref{tab:appendix_more_data}. The results show that even with the addition of large amounts of training data, existing methods still show limitations to the nighttime data that is obviously out-of-distribution from the training set. However, integrating FastTCM-CR50 can still perform robustly in such cases, indicating its robust generalization ability.

\begin{table}[ht]
\centering
\caption{Ablation study of exploration on large amounts of training data.
}	
\setlength\tabcolsep{1.0pt}
\begin{tabular}{lllll}
\toprule
Method       &BB  & Training Data       & Testing Data             & F (\%)      \\ 
\midrule
DB        &R50  & Joint data & NightTime-ArT   & 52.8        \\ 
DB        &CR50  & Joint data & NightTime-ArT   & 58.4        \\ \midrule
\multirow{2}[1]{*}{DB} &TCM-CR50 & Joint data  & NightTime-ArT  & {70.2}  \\ 
 &FastTCM-CR50 & Joint data  & NightTime-ArT  & \textbf{72.6}  \\ \bottomrule
\end{tabular}
	
\label{tab:appendix_more_data}
\end{table}

\subsection{Rotated Object Detection}
\label{subsec:rod}
To further validate the generalization ability of our approach, we adapted it to oriented object detection and evaluated its performance on the widely used DOTA-v1.0~\cite{Xia2017DOTAAL} dataset, which is specifically designed for oriented object detection in aerial images. The DOTA-v1.0 dataset consists of 15 common categories, 2806 images, and 188,282 instances. During training, we employed the same configuration as the cooperative methods for rotated object detection. 
T
As presented in Tab.~\ref{tab:dota}, we combined our model with previous approaches for oriented object detection. The results illustrate the consistent improvement by using the proposed FastTCM-CR50 backbone. 
We guess that the improvement of FastTCM-CR50  originates from its ability to utilize the rich prior knowledge offered by CLIP, thus optimizing the spotting and location of specific categories within satellite images. Specifically, FastTCM-CR50 initiates a synergy between visual features and their textual descriptions. Visual features aligning with textual descriptors are amplified, enabling the visual features to focus more on the segments related to remote sensing categories, thereby augmenting the performance of rotated object detection.

\begin{figure}[ht]
\centering
\includegraphics[width=0.48\textwidth]{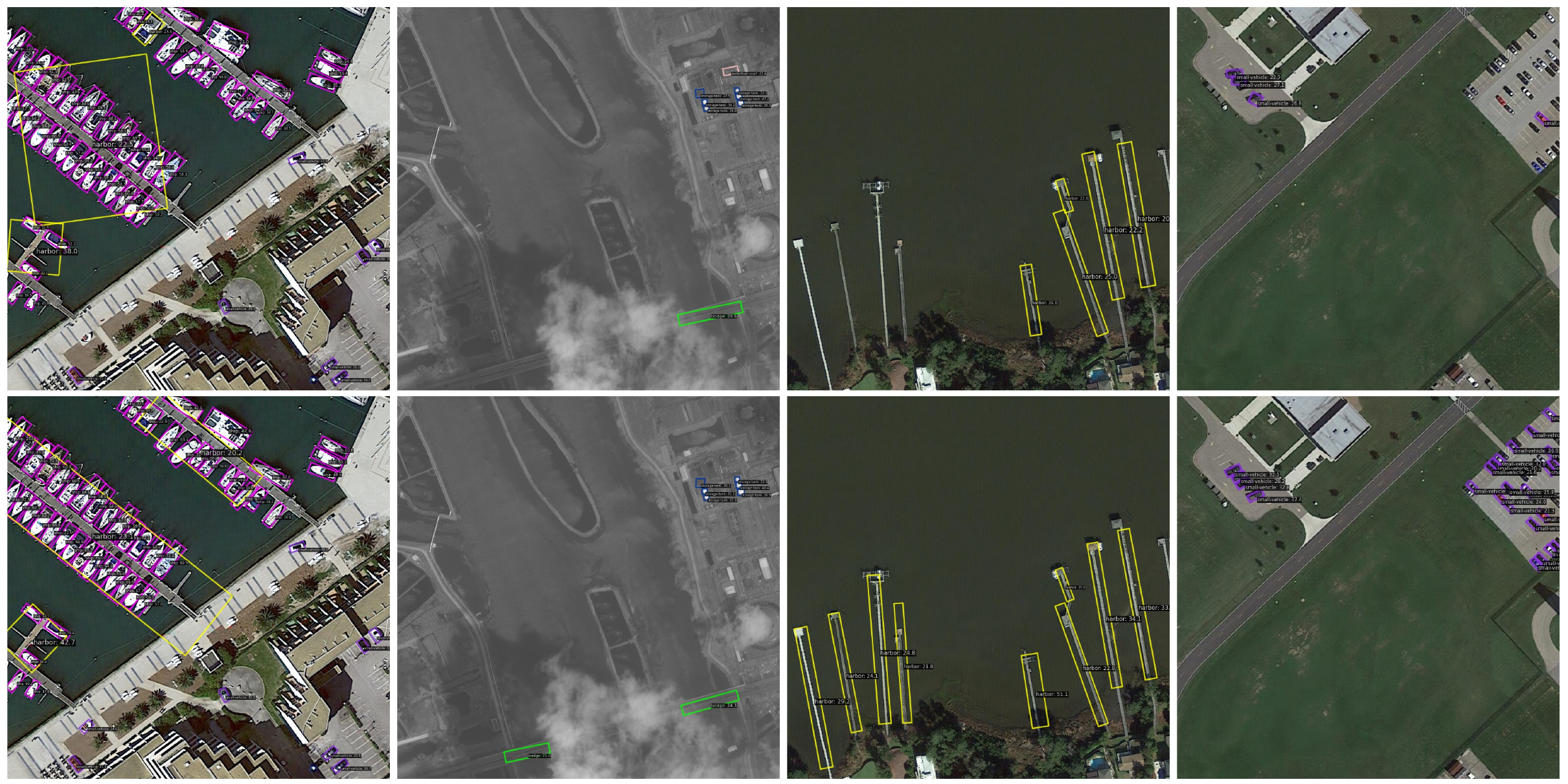}
\caption{The top row and bottom row are qualitative results on DOTA-v1.0 testing set without and with cooperating with FastTCM-CR50, respectively. It contains 15 common categories, such as ship, small-vehicle, harbor, bridge, basketball-court, storage-tank, etc. %
}
\label{fig:dota_vis}
\end{figure}

\begin{figure}[htbp]
\centering
\includegraphics[width=0.45\textwidth]{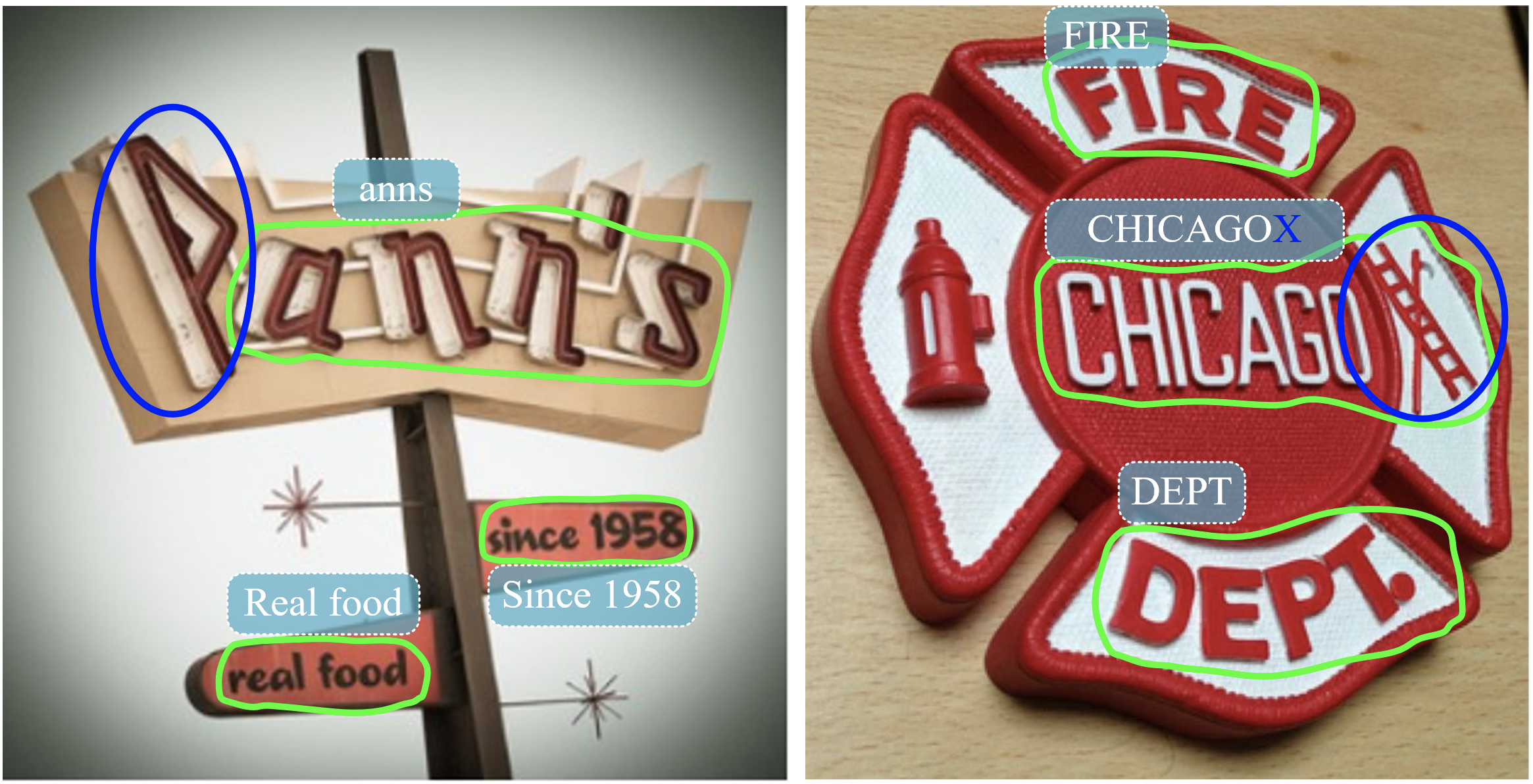}
\caption{Failure cases. Green polygons represent predicted detection results, and blue circle represents error detection regions. The dashed boxes stand for predicted recognition results, and blue characters are error recognition results.}
\label{fig:failure}
\end{figure}

\subsection{Summary of the Experiments}
The experimental analysis of FastTCM-CR50 in scene text detection and scene text spotting across various benchmarks demonstrates several advantages: (1) FastTCM can be seamlessly integrated to enhance existing scene text detectors and spotters with high efficiency. (2) FastTCM significantly improves the few-shot training ability of the detectors and spotters. (3) FastTCM also shows powerful generalization ability for generalization tasks, including domain adaptation, NightTime-ArT dataset, and rotated object detection dataset DOTA-v1.0. Some of the failure cases can be visualized in Fig. \ref{fig:failure}. We can see that some text-like objects might be mistakenly regarded as the positive text region.

\section{Conclusion}\label{sec:conclusion}
 The proposed FastTCM-CR50 backbone provides a notable enhancement to numerous scene text detectors and spotters, achieving consistent performance improvements, along with a significant increase in inference speed of 48.5\% compared to previous TCM-CR50. We conduct comprehensive ablation studies to demonstrate the effectiveness  every aspect of the proposed method. The robustness of FastTCM-CR50 is also demonstrated by its remarkable few-shot learning capabilities and generalization ability. Significant improvements on the NightTime-ArT subset from ICDAR2019-ArT and the rotated object detection dataset (DOTA-v1.0) further highlight the potential of the proposed method. We hope this work can provide a foundation for future advancements in the field of scene text detection and spotting.

\ifCLASSOPTIONcompsoc
  \section*{Acknowledgments}
\else
  \section*{Acknowledgment}
\fi

This work was supported by the National Natural Science Foundation of China (No.62225603, No.62206103).

\ifCLASSOPTIONcaptionsoff
  \newpage
\fi

{
\bibliographystyle{IEEEtran}
 \bibliography{strings_abbr, references}

}

\clearpage
\newpage
\appendix

\textbf{Dataset.}

\noindent    \textbf{ICDAR2013}~\cite{Karatzas2013ICDAR2R} is a high-resolution English dataset for focused scene text detection, including 229 images for training and 233 images for testing.
    
 \noindent   \textbf{ICDAR2015}~\cite{karatzas2015icdar} is a multi-oriented text detection dataset for English text that includes 1,000 training images and 500 testing images. Scene text images in this dataset were taken by Google Glasses without taking care of positioning, image quality, and viewpoint.
    
 \noindent   \textbf{MSRA-TD500}~\cite{yao2012detecting} is a multi-language dataset that includes English and Chinese, including 300 training images and 200 testing images. We also include extra 400 training images from HUST-TR400~\cite{Yao2014AUF} following the previous methods~\cite{Liao2020RealtimeST,Zhou2017EASTAE}.
    
\noindent    \textbf{CTW1500}~\cite{liu2019curved} consists of 1,000 training images and 500 testing images which focus on the curved text. The text instances are annotated at the text-line level by polygons with 14 vertices.
    
\noindent    \textbf{Total-Text}~\cite{ch2019total} contains 1,255 training images and 300 testing images. The text instances are labeled at the word level. It includes horizontal, multi-oriented, and curved text shapes.
    
\noindent    \textbf{ArT}~\cite{chng2019icdar2019} includes 5,603 training images and 4,563 testing images. It is a large-scale multi-lingual arbitrary-shape scene text detection dataset. The text regions are annotated by the polygons with an adaptive number of key points. Note that it contains Total-Text and CTW1500.
    
 \noindent   \textbf{MLT17}~\cite{Nayef2017ICDAR2017RR} includes 9 languages text representing 6 different scripts annotated by quadrangle. It has 7,200 training images, 1,800 validation images, and 9,000 testing images. We use both the training set and the validation set in the finetune period.
    
\noindent    \textbf{MLT19}~\cite{nayef2019icdar2019} is a large-scale multi-lingual scene text detection datasets. It contains 10,000 training images and 10,000 testing images and is labeled at the word level.
    
\noindent    \textbf{SynthText}~\cite{gupta2016synthetic} It contains 800k synthetic images generated by blending natural images with artificial text, which are all word-level annotated.
    
\noindent    \textbf{CurvedSynthText-150k} dataset~\cite{Liu2020ABCNetRS} comprises 150k assorted natural images overlaid with synthetically generated text, which incorporates both curved and regular typography for added diversity.
    
\noindent    \textbf{TextOCR} dataset~\cite{Singh2021TextOCRTL} is a large-scale, comprehensive, high-quality compilation of scene text images sourced from Open Images\footnote{\href{https://storage.googleapis.com/openimages/web/index.html}{Open Images Link}}. Averaging around 30 words per image, the dataset is characterized by extensive diversity. The dataset is divided into 24,902 images for training and 3,232 images for testing. Furthermore, each image is annotated with polygons for precise localization.

\begin{figure}[htbp]
\centering

\includegraphics[width=0.49\textwidth]{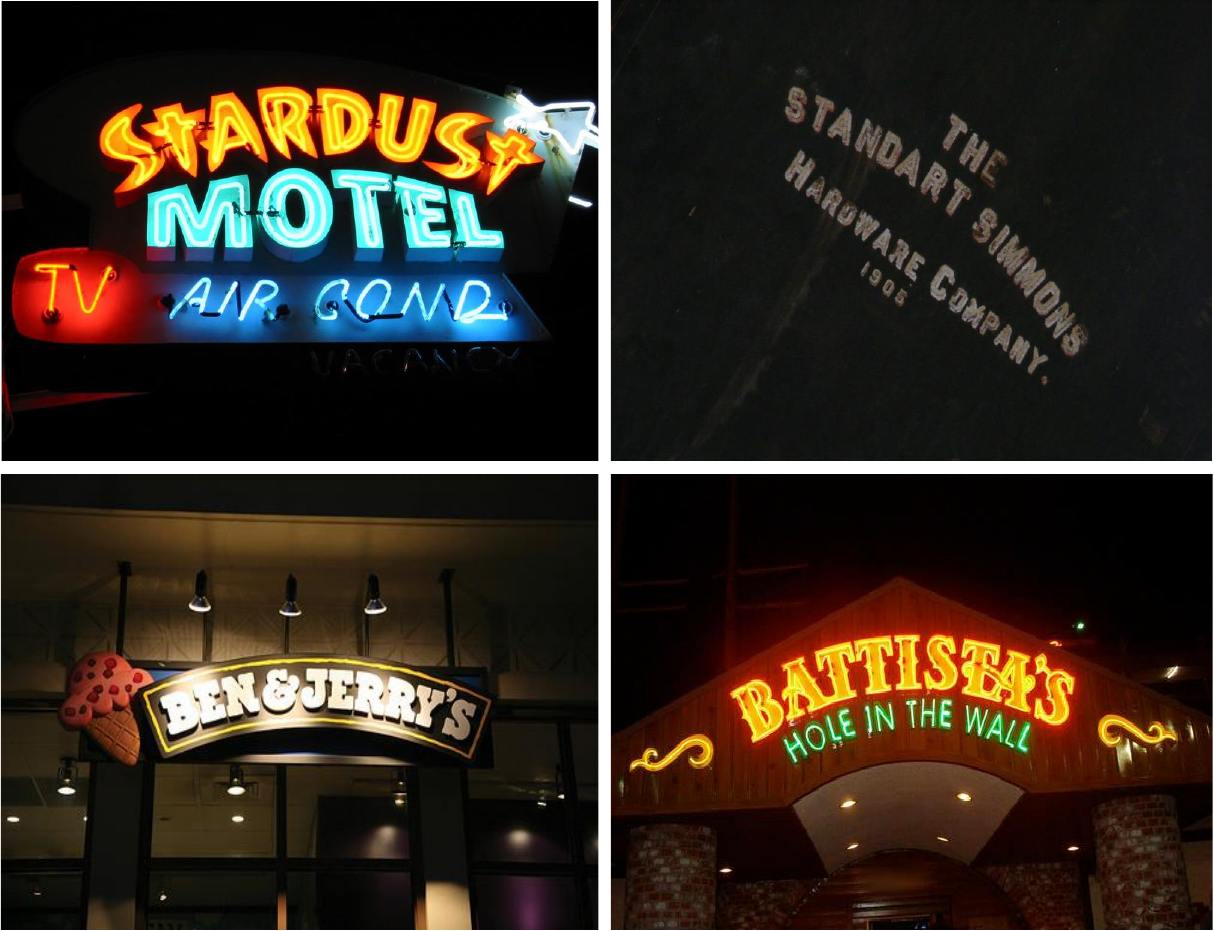}
\caption{The examples of our constructed NightTime-ArT.}
\label{fig:night_image}
\end{figure}

\noindent \textbf{NightTime-ArT Dataset} The NightTime-ArT dataset (326 images) carefully collected from ArT. The nighttime examples of ArT are provided in Fig.~\ref{fig:night_image}.

\noindent \textbf{Evaluation Metric.} For task detection tasks, we use intersection over union (IoU) to determine whether the model correctly detects the region of text, and we calculate precision (P), recall (R), and F-measure (F) for comparison following common practice~\cite{Karatzas2013ICDAR2R}. For fair comparisons, text regions labeled with either ``do not care'' or ``\#\#\#'' will be ignored in all datasets during training and testing. For end-to-end text spotting tasks, we follow the classic evaluation protocols and adopt H-mean as the primary metric.

\noindent \textbf{More Results.} \noindent\textbf{Ablation Study for Image Encoder and Text Encoder.} We have investigated how the quality of the frozen text encoder and image encoder affects the performance via adjusting the corresponding learning rate (LR) factor. The experimental results of DBNet with FastTCM-CR50 on the TD500 dataset are shown in Table~\ref{tab:appendix_expl_ie_te}. The results show that using a lower learning rate for both encoders and fixing the text encoder is the optimal setting for training the whole model. Note that we observe performance degradation when directly using $1.0\times$ learning rate for both encoders, which suggests the frozen text encoder can stabilize the training process. The cores of the architecture, including the language prompt generator and visual prompt generator, are designed to better steer knowledge of the pretrained CLIP. Appropriate design of the network architecture and the use of the pretrained CLIP are complementary.

	\begin{table}[ht]
		\centering
  		\caption{Ablation study of exploration on image encoder and text encoder with DBNet on TD dataset. 
		``LR'' represents the learning rate. F-measure (\%) is reported.
		}
		\renewcommand\arraystretch{1}

		\setlength{\tabcolsep}{2.0mm}

		\begin{tabular}{llll}
    \toprule
                           & Image encoder & Text encoder & TD \\ 
       \midrule
\multirow{4}{*}{LR Factor} & 0.1           & 0.0          & \textbf{88.9}  \\
                           & 0.1           & 0.1          & 87.9  \\
                           & 0.1           & 1.0          & 87.4  \\
                           & 1.0           & 1.0          & 86.5  \\ 
       \bottomrule
\end{tabular}

		\label{tab:appendix_expl_ie_te}
	\end{table}

\vfill 
\end{document}

%% file: table/text_det_ic15_td_ctw.tex
\newcommand{\tabincell}[2]{\begin{tabular}{@{}#1@{}}#2\end{tabular}}
{
\begin{tabularx}{1.0\linewidth}{ccccccccc}
    \toprule
	\multirow{2}[2]{*}{Method} & 
        \multirow{2}[2]{*}{BB} & 
	\multicolumn{2}{c}{IC15} &
	\multicolumn{2}{c}{TD} &
	\multicolumn{2}{c}{CTW} &
	\multirow{2}[1]{*}{FPS} 
	 \\
    \cmidrule(r){3-8}
	  & & F & $\Delta$ &  F & $\Delta$ & F & $\Delta$\\
    \midrule

    	TextSnake~\cite{Long2018TextSnakeAF} &R50 & 82.6 & - &  78.3 & - & 75.6 & - & 1.1 \\
	TextField~\cite{Xu2019TextFieldLA} &VGG16  & 84.1 & - & 81.3 & - & 81.4 & - &- \\
	PSENet-1s~\cite{Li2019ShapeRT}&R50  & 85.7 & -  & - & - & 82.2 & -  & 1.6\\
	LOMO~\cite{Zhang2019LookMT} &R50 & 87.2 & -  &  - & - & 78.4 & -  &-\\
	CRAFT~\cite{Baek2019CharacterRA} &VGG16 & 86.9 & -   & 82.9 & - & 83.5 & -& 8.6 \\
	ContourNet~\cite{Wang2020ContourNetTA} &R50  & 86.9   & - & - & - & 83.9 & -&- \\
	DRRG~\cite{Zhang2020DeepRR}  &VGG16 & 86.6 & -   & 85.1 & - & 84.5 & - &-\\
	MOST~\cite{He2021MOSTAM} &R50 & 88.2 & -  & 86.4 & - & - & - & 10.1 \\
	Raisi \etal\cite{Raisi2021TransformerbasedTD} &R50 & 83.7  & - & 87.2 & - & - & - &-\\
	Tang \etal\cite{Tang2022FewCB}  &R50 & 90.9 &- & 87.3 &-  & 89.1 & - & -\\
   TextFuseNet \cite{Ye2020TextFuseNetST}    &R50    &   90.1  &-   & - &- &  85.4 &- &- \\
   DB++ \cite{Liao2019RealTimeST}   &  R50  &  87.3 &-   &  87.2 &-   &   85.3 &- & 27\\
	TextBPN~\cite{Zhang2021AdaptiveBP} &R50 & - & -  & 85.6 & - & 85.0 & - &-\\

    \midrule

    	\multirow{4}[1]{*}{FCE\cite{zhu2021fourier}}  & R50& 86.2 & - & 85.4$^\dagger$ & - & 85.5 & - & 11.5 \\
 	 &CR50  & 86.2 & \textbf{\greenp{+0.2}}   & 86.1 & \textbf{\greenp{+0.7}} & 85.5 & \textbf{\greenp{+0.1}} & 11.5 \\
 \cline{2-9}
  	 &TCM-CR50  & 87.1 & \textbf{\greenp{+0.9}}   & 86.9 & \textbf{\greenp{+1.5}} & 85.9 & \textbf{\greenp{+0.4}} & 8.4 \\	
 	 &FastTCM-CR50  & 87.3 & \textbf{\greenp{+1.1}}   & 87.1 & \textbf{\greenp{+1.7}} & 86.0 & \textbf{\greenp{+0.5}} & 10.3 \\	
    \midrule
    
 	\multirow{4}[1]{*}{PAN\cite{Wang2019EfficientAA}}  &R50  & 82.9 & - & 84.1 & - & 83.7 & - & 36 \\
	 & CR50  & 83.2 & \textbf{\greenp{+0.3}} & 84.6 & \textbf{\greenp{+0.5}} & 83.9 & \textbf{\greenp{+0.2}} & 36 \\
 \cline{2-9}
 	 & TCM-CR50  & 84.6 & \textbf{\greenp{+1.7}} & {85.3} & \textbf{\greenp{+1.2}} & {84.3} & \textbf{\greenp{+0.6}} & 18 \\
   &FastTCM-CR50  & 84.9 & \textbf{\greenp{+2.0}} & {85.4} & \textbf{\greenp{+1.3}} & {84.5} & \textbf{\greenp{+0.8}} & 31 \\
    \cline{1-9}
    
 	\multirow{4}[1]{*}{DB\cite{Liao2020RealtimeST}}   &R50 & 87.3 & - & 84.9 & - & 83.4 & - & 14.5 \\
	 & CR50 & {87.7} & \textbf{\greenp{+0.4}} & {86.8} & \textbf{\greenp{+1.9}} & {83.4} & \textbf{\greenp{+0}} & 14.5  \\
 \cline{2-9}
	 & TCM-CR50 & {89.2} & \textbf{\greenp{+1.9}} & {88.8} & \textbf{\greenp{+3.9}} & {84.9} & \textbf{\greenp{+1.5}} & 10  \\
	 & FastTCM-CR50 & {89.4} & \textbf{\greenp{+2.1}} & \textbf{88.9} & \textbf{\greenp{+4.0}} & {85.2} & \textbf{\greenp{+1.8}} & 13.3  \\ 

    \bottomrule

\end{tabularx}
}

%% file: table/text_spot_and_det_only_tt_ic15_ctw_td.tex
\newcommand{\tabincell}[2]{\begin{tabular}{@{}#1@{}}#2\end{tabular}}
\begin{tabular}{lccccccccccccccccccccccccc}
    \toprule
	\multirow{3}[4]{*}{Method} & 
	\multirow{3}[4]{*}{BB} &
	\multirow{3}[4]{*}{Ext.} & 
	\multicolumn{6}{c}{Detection Results}  & \multicolumn{10}{c}{End-to-End Results}   & \multirow{3}[4]{*}{FPS}  \\
 \cmidrule(r){4-9}   \cmidrule(r){10-19}
 & & & 
	\multicolumn{2}{c}{TT} & 
        \multicolumn{2}{c}{IC15} &
        \multicolumn{2}{c}{CTW} &

        \multicolumn{3}{c}{TT} & 
        \multicolumn{4}{c}{IC15} &
        \multicolumn{3}{c}{CTW}

	 \\
    \cmidrule(r){4-5} \cmidrule(r){6-7} \cmidrule(r){8-9} \cmidrule(r){10-12} \cmidrule(r){13-16} \cmidrule(r){17-19} 
	&  & & F & $\Delta$ & F & $\Delta$ & F & $\Delta$   & None & Full & $\Delta$ (None) & S & W & G & $\Delta$ (S) & None & Full & $\Delta$ (None) \\

\midrule

	CharNet~\cite{Xing2019ConvolutionalCN} & R50& - &84.6 &-   &89.7 &- &- &-     &48.8 & 78.8 &-   &80.1 &74.5 &62.2 &-  &- &-   &-  &5.4   \\  
      CRAFTS~\cite{Baek2020CharacterRA} &R50 & -  & 87.4 &-    &87.1 &- &- &-       &78.8 &- &-   &83.1 &82.1 &74.9 &- &- &- &-   &8.8      \\
	TextPerceptron~\cite{Qiao2020TextPT} & R50& - & 85.2 &-   &87.1 &-  &84.6 &- &    69.7 & 78.3 &-  &80.1 & 76.6 &65.1 &-  &57.0 &- &-  &- \\
 
	TextDragon\cite{Feng2019TextDragonAE}  & VGG16 & - & 80.3 &- & 87.8 &- &83.6  &-    & 48.8  & 74.8 &- &82.5  &78.3 &65.2 &-   &39.7 &72.4 &- &2.6 \\	
	Boundary\cite{Wang2020AllYN} & R50 & -  &87.0 &- &88.6 &- &- &-    &65.0 &76.1 &- &79.7 &75.2 &64.1 &- &- &- &- &- \\	
	PAN++\cite{Wang2021PANTE} & R18 & - &- &-   &87.5 &-  &84.0 &- & 68.6 &78.6&- &82.7 &78.2 &69.2 &- &- &- &- &36.0 \\	
	PGNet\cite{Wang2021PGNetRA} & R50  & -  &86.1&- &88.2 &-  &- &-  &63.1 &- &-  &88.3 &78.3 &63.5 &-  &- &- &- &38.2 \\	
 	MANGO\cite{Qiao2020MANGOAM}  & R50 & -  &- &- &- &-  &- &-  &72.9 &83.6&- &85.4 &80.1 &73.9 &- &58.9 &78.7 &-  & 8.4 \\

        GLASS~\cite{Ronen2022GLASSGT}& R50  & -   &88.1 &- &85.7 &- &- &-    &79.9 &86.2 &- &84.7 &80.1 &76.3 &- &- &- &- &2.7  \\	
        MTSv2~\cite{liao2019mask} & R50 &- & 78.5 &- &87.0 &- &- &- &65.3 &77.4 &- &83.0 &77.7 &73.5 &- &- &- &- &3.1  \\
        ABCNetv2~\cite{Liu2021ABCNetVA} & R50 &-  &87.0 &- &88.1 &- &84.7 &- &73.5 &80.7 &- &83.0 &80.7 &75.0 &- &58.4 &79.0 &- & 10 \\
        SwinTS~\cite{Huang2022SwinTextSpotterST} & Swin-T & -  &88.0 &-  &- &- &88.0 &-  &74.3 &84.1 & -  &83.9 &77.3 &70.5 &- &51.8 &77.0 &-  &2.9  \\	
        SPTS\cite{Peng2021SPTSST} & R50 & -  &- &-  & - &- &- &-   &74.2 &82.4 &- &77.5 &70.2 &65.8 &- &63.6 &83.8 &- &0.6\\		
        TTS~\cite{Kittenplon2022TowardsWT} & R50 & -  &- &-  &- &- &- &-  &78.2 &86.3 &-  &85.2 &81.7 &77.4 &-  &- &- &-  &- \\	

\midrule
\multirow{4}[1]{*}{MTSv3~\cite{Liao2020MaskTV}}  & R50 & -  & 79.7 & - & 87.5 &- & 83.7 &   -  &71.2 &78.4 &  -  &83.3 &78.1 &74.2 & -  &52.6 &75.8 &-  & 2.5 \\
 & oCLIP-R50  &- & 80.0 & \textbf{\greenp{+0.3}}  &87.7 &\textbf{\greenp{+0.2}}  &83.9 &\textbf{\greenp{+0.2}}   & 71.8  & 79.6 &\textbf{\greenp{+0.6}}   & 83.8 & 78.6 & 74.3 & \textbf{\greenp{+0.5}} &53.0 &76.4 &\textbf{\greenp{+0.4}}  &2.5 \\
 & TCM-CR50   & - & 81.5 & \textbf{\greenp{+1.8}}  & 88.0 &\textbf{\greenp{+0.5}}  &84.1 &\textbf{\greenp{+0.4}}   & 73.7 & 82.6& \textbf{\greenp{+2.5}} & 84.1 & 80.2& 75.5& \textbf{\greenp{+0.8}}   &53.2 &76.7 &\textbf{\greenp{+0.6}} & 1.3\\
  & FastTCM-CR50   & - & 81.9 & \textbf{\greenp{+2.2}}  & 88.2 &\textbf{\greenp{+0.7}}  & 84.2 &\textbf{\greenp{+0.5}}  & 73.9 & 83.0 & \textbf{\greenp{+2.7}}  & 84.5 & 80.3 & 76.0 & \textbf{\greenp{+1.2}} & 54.0 &77.2 & \textbf{\greenp{+1.4}} & 1.9 \\
	
\midrule
\multirow{3}[2]{*}{ABINet++~\cite{Fang2022ABINetAB}} & R50 & -   & 86.0 & -  & 88.2 & -  & 84.0  &   -  & 77.6  & 84.5 &   -  &84.1 &80.4 &75.4 & -   &60.2 & 80.3 &-&10.6\\
 & TCM-CR50 & -  & 86.3 &  \textbf{\greenp{+0.3}} & 88.4 & \textbf{\greenp{+0.2}}  & 84.3  & \textbf{\greenp{+0.3}}& 77.7  &84.8  &\textbf{\greenp{+0.1}}  &84.4 &80.7 &75.7 &\textbf{\greenp{+0.3}}       &60.4 &80.5  &\textbf{\greenp{+0.2}} &5.1\\
   & FastTCM-CR50 & -  & 86.5 &  \textbf{\greenp{+0.5}} & 88.8 & \textbf{\greenp{+0.6}}  & 84.7  &\textbf{\greenp{+0.7}}    & 77.9  & 85.0 & \textbf{\greenp{+0.3}} &85.1 &80.9 &76.0 & \textbf{\greenp{+1.0}}    &60.5 &80.8  & \textbf{\greenp{+0.3}} &9.7\\	
 \midrule
     
 \multirow{3}[2]{*}{ABCNet~\cite{Liu2020ABCNetRS}}    & R50 & -   & 86.0 & -  & 86.8 & - &84.4 &-  & 64.2 & 75.7 & - & 79.2  & 74.1 & 66.8 & -    & 45.2 & 74.1 &-&17.9\\	
      & CR50 & -  & 86.1 & \textbf{\greenp{+0.1}} & 87.2 & \textbf{\greenp{+0.4}} & 84.5 &   \textbf{\greenp{+0.1}} & 64.3 &76.4 & \textbf{\greenp{+0.1}} &79.4 & 75.0 & 67.3 & \textbf{\greenp{+0.2}} &46.1 & 75.3 & \textbf{\greenp{+0.9}} & 17.9\\
	 & TCM-CR50 & -  & 86.4 & \textbf{\greenp{+0.4}} & 88.4 & \textbf{\greenp{+1.6}} &84.8 &   \textbf{\greenp{+0.4}}  & 68.3 &  81.8 & \textbf{\greenp{+4.1}} & 81.5  & 77.2 &  72.3 & \textbf{\greenp{+2.3}}  &48.2 & 74.6 & \textbf{\greenp{+3.0}}  &16.4\\
     & FastTCM-CR50  & - & 86.6 & \textbf{\greenp{+0.7}}  & 88.9 & \textbf{\greenp{+2.1}}  &85.3 & \textbf{\greenp{+0.9}}  & 69.5 &  82.4 & \textbf{\greenp{+5.3}} &  82.8  & 78.1  &  72.8 & \textbf{\greenp{+3.6}}  & 49.1 & 77.8 &  \textbf{\greenp{+3.9}} &17.2\\
    \midrule
    
 \multirow{3}[2]{*}{DeepSolo\cite{Ye2022DeepSoloLT}}   & R50 & -  & 87.3 & - &90.0 &  -  &87.2  &-  & 79.7 & 87.0 &  -  & 86.8 &81.9 &76.9 & -   &60.1 & 78.4 &-&17.0  \\	
   & CR50 & -  & 87.4 &  \textbf{\greenp{+0.1}} & 90.1 & \textbf{\greenp{+0.1}}   & 87.2  &  \textbf{\greenp{+0}}  &79.7 & 87.0 & \textbf{\greenp{+0}}  & 86.8 & 81.9 &77.0 & \textbf{\greenp{+0}} &60.2 & 78.4 & \textbf{\greenp{+0.1}} & 17.0\\
  & TCM-CR50 & -  & 87.5 &  \textbf{\greenp{+0.2}} & 90.2 & \textbf{\greenp{+0.2}}   & 87.3  & \textbf{\greenp{+0.1}}  & 79.8 &  87.1&  \textbf{\greenp{+0.1}}  & 86.9  & 82.0 & 77.1 &  \textbf{\greenp{+0.1}}    & 60.3 &78.5  &\textbf{\greenp{+0.2}} &8.3\\
 & FastTCM-CR50 & -  & 87.8 & \textbf{\greenp{+0.5}}  & 90.3 &  \textbf{\greenp{+0.3}}   &87.4 &\textbf{\greenp{+0.2}}  & 79.9 & 87.2 &  \textbf{\greenp{+0.2}} &  87.0 &82.0 & 77.3 &  \textbf{\greenp{+0.2}}     & 60.4 & 78.8 & \textbf{\greenp{+0.4}} &15.1 \\

    \midrule

 \multirow{4}[2]{*}{ \tabincell{c}{TESTR~\cite{Zhang2022TextST} \\ (polygon)}} 	 & R50 & -  & 86.9 & - & 90.0 & - &87.1 &-  & 73.2 & 83.9 & -  &  85.2 &79.4& 73.6 & -     &55.9 &81.5 &-&12.1 \\
 	  & CR50 & - &  87.1 & \textbf{\greenp{+0.2}}  & 90.0 & \textbf{\greenp{+0}} & 87.1 & \textbf{\greenp{+0}}  &73.3 &84.0 & \textbf{\greenp{+0.1}} & 85.4 & 80.6 &74.6 & \textbf{\greenp{+0.3}} &56.1 & 81.6 & \textbf{\greenp{+0.2}} & 12.1\\
	  & TCM-CR50 & - & 87.9 & \textbf{\greenp{+1.0}}  & 90.1 & \textbf{\greenp{+0.1}} & 87.2 & \textbf{\greenp{+0.1}}   & 73.6 & 84.1 &  \textbf{\greenp{+0.4}} &  85.6 & 80.3 & 74.2 &  \textbf{\greenp{+0.4}}    &56.2 & 81.8 & \textbf{\greenp{+0.3}} &6.2 \\	
 	 & FastTCM-CR50 & -  & 88.1 & \textbf{\greenp{+1.2}} & 90.2 & \textbf{\greenp{+0.2}} & 87.3 &  \textbf{\greenp{+0.2}}   & 74.2 & 85.4 &  \textbf{\greenp{+1.0}} & 86.2 & 80.9 & 75.2 &\textbf{\greenp{+1.0}}    &56.8 & 82.4 & \textbf{\greenp{+0.9}}&10.7\\	
   	 & FastTCM-CR50 & \checkmark   & 89.0 & \textbf{\greenp{+2.1}} & 90.3 & \textbf{\greenp{+0.3}}  & 87.4 &\textbf{\greenp{+0.3}} & 76.5 & 86.9 &  \textbf{\greenp{+3.3}}  & 86.8 & 81.6 & 76.1 & \textbf{\greenp{+1.6}}  &57.5 & 82.8 & \textbf{\greenp{+1.6}}&10.7 \\	
    \bottomrule
\end{tabular}

%% file: table/synth_to_real.tex
\newcommand{\tabincell}[2]{\begin{tabular}{@{}#1@{}}#2\end{tabular}}
\begin{tabular}{lccc}
    \toprule
	\multirow{1}[0]{*}{Method} &
 	\multirow{1}[0]{*}{BB} &
	\multicolumn{1}{c}{ST $\rightarrow$ IC13} &
	\multicolumn{1}{c}{ST $\rightarrow$ IC15}
	 \\
    \midrule
	EAST$^\dagger$~\cite{Zhou2017EASTAE} &PVANet~\cite{Kim2016PVANETDB} & 67.1 & 60.5   \\
	PAN~\cite{Wang2019EfficientAA} &R50 & - & 54.8   \\
	CCN~\cite{Xing2019ConvolutionalCN} &R50 & - & 65.1  \\
	ST3D~\cite{Liao2020SynthText3DSS} &R50 & 73.8 & 67.6  \\
    \midrule
	\multirow{4}[2]{*}{DB~\cite{Liao2020RealtimeST}}  &R50 & 71.7 & 64.0  \\
           &    CR50      &  73.1    &   67.4     \\
	  & TCM-CR50   & {79.6} & {76.7}  \\
 	 & FastTCM-CR50 & \textbf{79.9} & \textbf{77.2}  \\
    \bottomrule
\end{tabular}

%% file: table/real_to_real.tex
\newcommand{\tabincell}[2]{\begin{tabular}{@{}#1@{}}#2\end{tabular}}
\begin{tabular}{lcccc}
    \toprule
	\multirow{1}[0]{*}{Method} &
 	\multirow{1}[0]{*}{BB} &
	\multicolumn{1}{c}{IC13$\rightarrow$IC15} &
	\multicolumn{1}{c}{IC13$\rightarrow$TD} &
 	\multicolumn{1}{c}{MLT17$\rightarrow$MLT19}
	 \\
    \midrule
	EAST$^\dagger$~\cite{Zhou2017EASTAE} &PVANet & 53.3 & 46.8 &-  \\
	GD(AD)~\cite{Zhan2019GADANGD} &- & 64.4  & 58.5  &-\\
	GD(10-AD)\cite{Zhan2019GADANGD} &-  & 69.4  & 62.1 &-  \\
	CycleGAN~\cite{Zhu2017UnpairedIT}  &- & 57.2  & -  &- \\
	ST-GAN~\cite{Lin2018STGANST} &- & 57.6  & - &-  \\
	TST~\cite{Wu2020SynthetictoRealUD} &PVANet & 52.4  & -  &- \\
     \midrule
	\multirow{3}[2]{*}{DB~\cite{Liao2020RealtimeST}}   &R50 & 63.9 & 53.8  &47.4\\
	 &TCM-CR50 & {71.9}  & {65.1}  &{67.5} \\
 	 & FastTCM-CR50 & \textbf{72.4}  & \textbf{65.7}  &\textbf{67.8} \\
    \bottomrule
\end{tabular}

%% file: table/ablation_ic15_td_tt_ctw.tex
\newcommand{\tabincell}[2]{\begin{tabular}{@{}#1@{}}#2\end{tabular}}
{
\begin{tabularx}{1.0\linewidth}{ccccccccccc}
\hline
    \toprule
	\multirow{2}[2]{*}{Method} & \multirow{2}[2]{*}{PP} &
	\multirow{2}[2]{*}{LP} &
	\multirow{2}[2]{*}{\tabincell{c}{LG}} & 
    \multirow{2}[2]{*}{\tabincell{c}{VG}} & 
        \multirow{2}[2]{*}{\tabincell{c}{Aux.}} & 
            \multirow{2}[2]{*}{\tabincell{c}{BSM}} & 
	\multicolumn{1}{c}{IC15} &
	\multicolumn{1}{c}{TD} &
	\multicolumn{1}{c}{TT} &
	\multicolumn{1}{c}{CTW} 
	 \\
    \cmidrule(r){8-11}
	& & & & & & &  F & F & F & F \\
    \midrule

	BSL & $\times$  & $\times$  & $\times$  & $\times$ & \checkmark & $\times$ & 87.7 & 86.8 & 84.7 & 83.4  \\
    \midrule

	\multirow{10}[1]{*}{BSL+} & \checkmark  & $\times$  & $\times$  & $\times$&  \checkmark& $\times$ & 87.75  & 87.0 & 84.74 &  83.5  \\
	   & \checkmark  & 4  & $\times$  & $\times$  &  \checkmark & $\times$ & 88.0  & 87.1 & 84.8 &  83.6   \\
       & $\times$   & 4  & $\times$  & $\times$ &  \checkmark & $\times$& 87.8  & 87.7 &  85.1 &  83.9   \\
	  & $\times$   & 18  & $\times$  & $\times$&  \checkmark  & $\times$  & 88.1  & 87.8 & 85.3 & 83.9   \\
	& $\times$   & 32  & $\times$  & $\times$ &  \checkmark & $\times$ & 88.4  &  88.2 & 85.4 &  84.5  \\
	 & \checkmark  & 4  & \checkmark  & $\times$& \checkmark & $\times$ &  88.6  & 88.4  & 85.5 &  84.6  \\
	 & \checkmark  & 4  & \checkmark  & \checkmark &  \checkmark & $\times$ & 89.2  & 88.9  & 85.6  & 84.9  \\
	 & \checkmark  & 32  & \checkmark  & \checkmark &  \checkmark & $\times$ & {89.4}   & {88.8}   & {85.9} & {85.1}   \\
      & \checkmark  & 4  & \checkmark  & \checkmark &   $\times$ &\checkmark  & 87.9   & 87.2  & 84.6 & 84.2  \\
      & \checkmark  & 4  & \checkmark  & \checkmark  &   \checkmark &  $\times$  & 88.1   & 87.5  & 85.0 & 84.6  \\
        \midrule
        FastTCM-CR50 & \checkmark  & 4  & \checkmark  & \checkmark  &   \checkmark &\checkmark  & \textbf{89.4}   & \textbf{88.9}   & \textbf{86.1} & \textbf{85.2}   \\
    \bottomrule
\end{tabularx}
}

%% file: table/abla_da_tg_vg.tex
\newcommand{\tabincell}[2]{\begin{tabular}{@{}#1@{}}#2\end{tabular}}
\begin{tabular}{lcccc}
    \toprule

    	\multirow{1}[0]{*}{Method} & 
 	\multirow{1}[0]{*}{BB} & 
	\multicolumn{1}{c}{TT $\rightarrow$ IC15} &
	\multicolumn{1}{c}{TT $\rightarrow$ CTW} &
	\multicolumn{1}{c}{IC15 $\rightarrow$ CTW} 
	 \\
  \midrule
  	TESTR & FastTCM-CR50  & \textbf{53.1} &  \textbf{49.2}  & \textbf{48.1}  \\
    \midrule

	\multirow{3}[1]{*}{TESTR} & w/o MQ   & 50.4 (\redp{-2.7})  & 45.5 (\redp{-3.7})   & 44.6 (\redp{3.5})  \\
	& w/o BSM   & 48.2 (\redp{-3.9})  & 43.4 (\redp{-5.8}) & 42.9 (\redp{-5.2})  \\
     & w/o LG    & 45.5 (\redp{-7.6}) & 40.3  (\redp{-9.0}) & 41.7 (\redp{-6.4})  \\
     & w/o VG    & 46.6 (\redp{-6.5}) & 42.4 (\redp{-6.8}) &  42.0 (\redp{-6.1}) \\
 
 & w/o All & 41.4 (\redp{-11.7}) & 39.2 (\redp{-10}) & 37.3 (\redp{-10.8})    \\	
    \bottomrule
\end{tabular}

%% file: arxiv.bbl
\begin{thebibliography}{10}
\providecommand{\url}[1]{#1}
\csname url@samestyle\endcsname
\providecommand{\newblock}{\relax}
\providecommand{\bibinfo}[2]{#2}
\providecommand{\BIBentrySTDinterwordspacing}{\spaceskip=0pt\relax}
\providecommand{\BIBentryALTinterwordstretchfactor}{4}
\providecommand{\BIBentryALTinterwordspacing}{\spaceskip=\fontdimen2\font plus
\BIBentryALTinterwordstretchfactor\fontdimen3\font minus
  \fontdimen4\font\relax}
\providecommand{\BIBforeignlanguage}[2]{{%
\expandafter\ifx\csname l@#1\endcsname\relax
\typeout{** WARNING: IEEEtran.bst: No hyphenation pattern has been}%
\typeout{** loaded for the language `#1'. Using the pattern for}%
\typeout{** the default language instead.}%
\else
\language=\csname l@#1\endcsname
\fi
#2}}
\providecommand{\BIBdecl}{\relax}
\BIBdecl

\bibitem{Radford2021LearningTV}
A.~Radford, J.~W. Kim, C.~Hallacy, A.~Ramesh, G.~Goh, S.~Agarwal, G.~Sastry,
  A.~Askell, P.~Mishkin, J.~Clark, G.~Krueger, and I.~Sutskever, ``Learning
  transferable visual models from natural language supervision,'' in
  \emph{ICML}, 2021, pp. 1--16.

\bibitem{Zhou2022ConditionalPL}
K.~Zhou, J.~Yang, C.~C. Loy, and Z.~Liu, ``Conditional prompt learning for
  vision-language models,'' in \emph{CVPR}, 2022, pp. 16\,816--16\,825.

\bibitem{Gu2022OpenvocabularyOD}
X.~Gu, T.-Y. Lin, W.~Kuo, and Y.~Cui, ``Open-vocabulary object detection via
  vision and language knowledge distillation,'' in \emph{ICLR}, 2022, pp.
  1--20.

\bibitem{Rao2022DenseCLIPLD}
Y.~Rao, W.~Zhao, G.~Chen, Y.~Tang, Z.~Zhu, G.~Huang, J.~Zhou, and J.~Lu,
  ``Denseclip: Language-guided dense prediction with context-aware prompting,''
  in \emph{CVPR}, 2022, pp. 18\,061--18\,070.

\bibitem{Xu2021ASB}
M.~Xu, Z.~Zhang, F.~Wei, Y.~Lin, Y.~Cao, H.~Hu, and X.~Bai, ``A simple baseline
  for zero-shot semantic segmentation with pre-trained vision-language model,''
  in \emph{ECCV}, 2021.

\bibitem{Song2022VisionLanguagePF}
S.~Song, J.~Wan, Z.~Yang, J.~Tang, W.~Cheng, X.~Bai, and C.~Yao,
  ``Vision-language pre-training for boosting scene text detectors,'' in
  \emph{CVPR}, 2022, pp. 15\,681--15\,691.

\bibitem{Wan2021SelfattentionBT}
Q.~Wan, H.~Ji, and L.~Shen, ``Self-attention based text knowledge mining for
  text detection,'' in \emph{CVPR}, 2021, pp. 5979--5988.

\bibitem{Xue2022LanguageMA}
C.~Xue, W.~Zhang, Y.~Hao, S.~Lu, P.~H.~S. Torr, and S.~Bai, ``Language matters:
  A weakly supervised vision-language pre-training approach for scene text
  detection and spotting,'' in \emph{ECCV}, 2022, pp. 1--19.

\bibitem{Yu2023TurningAC}
W.~Yu, Y.~Liu, W.~Hua, D.~Jiang, B.~Ren, and X.~Bai, ``Turning a clip model
  into a scene text detector,'' in \emph{CVPR}, 2023.

\bibitem{Shi2017DetectingOT}
B.~Shi, X.~Bai, and S.~J. Belongie, ``Detecting oriented text in natural images
  by linking segments,'' in \emph{CVPR}, 2017, pp. 3482--3490.

\bibitem{Long2018TextSnakeAF}
S.~Long, J.~Ruan, W.~Zhang, X.~He, W.~Wu, and C.~Yao, ``Textsnake: A flexible
  representation for detecting text of arbitrary shapes,'' in \emph{ECCV},
  2018, pp. 1--17.

\bibitem{Li2019ShapeRT}
X.~Li, W.~Wang, W.~Hou, R.-Z. Liu, T.~Lu, and J.~Yang, ``Shape robust text
  detection with progressive scale expansion network,'' in \emph{CVPR}, 2019,
  pp. 9328--9337.

\bibitem{Wang2019EfficientAA}
W.~Wang, E.~Xie, X.~Song, Y.~Zang, W.~Wang, T.~Lu, G.~Yu, and C.~Shen,
  ``Efficient and accurate arbitrary-shaped text detection with pixel
  aggregation network,'' in \emph{ICCV}, 2019, pp. 8439--8448.

\bibitem{Liao2020RealtimeST}
M.~Liao, Z.~Wan, C.~Yao, K.~Chen, and X.~Bai, ``Real-time scene text detection
  with differentiable binarization,'' in \emph{AAAI}, 2020, pp.
  11\,474--11\,481.

\bibitem{Tang2022FewCB}
J.~R. Tang, W.~Zhang, H.~yi~Liu, M.~Yang, B.~Jiang, G.-N. Hu, and X.~Bai, ``Few
  could be better than all: Feature sampling and grouping for scene text
  detection,'' in \emph{CVPR}, 2022, pp. 4563--4572.

\bibitem{Long2022TowardsEU}
S.~Long, S.~Qin, D.~Panteleev, A.~Bissacco, Y.~Fujii, and M.~Raptis, ``Towards
  end-to-end unified scene text detection and layout analysis,'' in
  \emph{CVPR}, 2022, pp. 1039--1049.

\bibitem{Zhang2016MultiorientedTD}
Z.~Zhang, C.~Zhang, W.~Shen, C.~Yao, W.~Liu, and X.~Bai, ``Multi-oriented text
  detection with fully convolutional networks,'' in \emph{CVPR}, 2016, pp.
  4159--4167.

\bibitem{Liu2017DeepMP}
Y.~Liu and L.~Jin, ``Deep matching prior network: Toward tighter multi-oriented
  text detection,'' in \emph{CVPR}, 2017, pp. 3454--3461.

\bibitem{He2017SingleST}
P.~He, W.~Huang, T.~He, Q.~Zhu, Y.~Qiao, and X.~Li, ``Single shot text detector
  with regional attention,'' in \emph{ICCV}, 2017, pp. 3066--3074.

\bibitem{He2017DeepDR}
W.~He, X.-Y. Zhang, F.~Yin, and C.-L. Liu, ``Deep direct regression for
  multi-oriented scene text detection,'' in \emph{ICCV}, 2017, pp. 745--753.

\bibitem{Liao2017TextBoxesAF}
M.~Liao, B.~Shi, X.~Bai, X.~Wang, and W.~Liu, ``Textboxes: A fast text detector
  with a single deep neural network,'' in \emph{AAAI}, 2017, pp. 4161--4167.

\bibitem{Zhou2017EASTAE}
X.~Zhou, C.~Yao, H.~Wen, Y.~Wang, S.~Zhou, W.~He, and J.~Liang, ``East: An
  efficient and accurate scene text detector,'' in \emph{CVPR}, 2017, pp.
  2642--2651.

\bibitem{Zhang2019LookMT}
C.~Zhang, B.~Liang, Z.~Huang, M.~En, J.~Han, E.~Ding, and X.~Ding, ``Look more
  than once: An accurate detector for text of arbitrary shapes,'' in
  \emph{CVPR}, 2019, pp. 10\,544--10\,553.

\bibitem{Wang2019ArbitrarySS}
X.~Wang, Y.~Jiang, Z.~Luo, C.-L. Liu, H.~Choi, and S.~Kim, ``Arbitrary shape
  scene text detection with adaptive text region representation,'' in
  \emph{CVPR}, 2019, pp. 6442--6451.

\bibitem{zhu2021fourier}
Y.~Zhu, J.~Chen, L.~Liang, Z.~Kuang, L.~Jin, and W.~Zhang, ``Fourier contour
  embedding for arbitrary-shaped text detection,'' in \emph{CVPR}, 2021, pp.
  3122--3130.

\bibitem{He2021MOSTAM}
M.~He, M.~Liao, Z.~Yang, H.~Zhong, J.~Tang, W.~Cheng, C.~Yao, Y.~Wang, and
  X.~Bai, ``Most: A multi-oriented scene text detector with localization
  refinement,'' in \emph{CVPR}, 2021, pp. 8809--8818.

\bibitem{Zhang2021AdaptiveBP}
S.-X. Zhang, X.~Zhu, C.~Yang, H.~Wang, and X.-C. Yin, ``Adaptive boundary
  proposal network for arbitrary shape text detection,'' in \emph{ICCV}, 2021,
  pp. 1285--1294.

\bibitem{Dai2021ProgressiveCR}
P.~Dai, S.~Zhang, H.~Zhang, and X.~Cao, ``Progressive contour regression for
  arbitrary-shape scene text detection,'' in \emph{CVPR}, 2021, pp. 7389--7398.

\bibitem{Ye2022DPTextDETRTB}
M.~Ye, J.~Zhang, S.~Zhao, J.~Liu, B.~Du, and D.~Tao, ``Dptext-detr: Towards
  better scene text detection with dynamic points in transformer,'' in
  \emph{AAAI}, 2023.

\bibitem{Zhang2022ArbitraryST}
S.-X. Zhang, X.~Zhu, C.~Yang, and X.-C. Yin, ``Arbitrary shape text detection
  via boundary transformer,'' \emph{TMM}, 2023.

\bibitem{li2017towards}
H.~Li, P.~Wang, and C.~Shen, ``{Towards end-to-end text spotting with
  convolutional recurrent neural networks},'' in \emph{ICCV}, 2017, pp.
  5238--5246.

\bibitem{ren2015faster}
S.~Ren, K.~He, R.~Girshick, and J.~Sun, ``Faster r-cnn: Towards real-time
  object detection with region proposal networks,'' in \emph{NeurIPS}, 2015,
  pp. 91--99.

\bibitem{li2019towards}
H.~Li, P.~Wang, and C.~Shen, ``Towards end-to-end text spotting in natural
  scenes,'' \emph{TPAMI}, 2019.

\bibitem{busta2017deep}
M.~Busta, L.~Neumann, and J.~Matas, ``Deep textspotter: An end-to-end trainable
  scene text localization and recognition framework,'' in \emph{ICCV}, 2017,
  pp. 2204--2212.

\bibitem{he2018end}
T.~He, Z.~Tian, W.~Huang, C.~Shen, Y.~Qiao, and C.~Sun, ``An end-to-end
  textspotter with explicit alignment and attention,'' in \emph{CVPR}, 2018,
  pp. 5020--5029.

\bibitem{liu2018fots}
X.~Liu, D.~Liang, S.~Yan, D.~Chen, Y.~Qiao, and J.~Yan, ``{Fots: Fast oriented
  text spotting with a unified network},'' in \emph{CVPR}, 2018, pp.
  5676--5685.

\bibitem{lyu2018mask}
P.~Lyu, M.~Liao, C.~Yao, W.~Wu, and X.~Bai, ``Mask textspotter: An end-to-end
  trainable neural network for spotting text with arbitrary shapes,'' in
  \emph{ECCV}, 2018, pp. 67--83.

\bibitem{liao2019mask}
M.~Liao, P.~Lyu, M.~He, C.~Yao, W.~Wu, and X.~Bai, ``Mask textspotter: An
  end-to-end trainable neural network for spotting text with arbitrary
  shapes,'' \emph{TPAMI}, vol.~43, no.~2, pp. 532--548, 2021.

\bibitem{qin2019towards}
S.~Qin, A.~Bissacco, M.~Raptis, Y.~Fujii, and Y.~Xiao, ``Towards unconstrained
  end-to-end text spotting,'' in \emph{ICCV}, 2019.

\bibitem{feng2019textdragon}
F.~Wei, H.~Wenhao, Y.~Fei, Z.~Xu-Yao, and C.-L. Liu, ``{TextDragon}: An
  end-to-end framework for arbitrary shaped text spotting,'' in \emph{ICCV},
  2019.

\bibitem{Wang2020AllYN}
H.~Wang, P.~Lu, H.~Zhang, M.~Yang, X.~Bai, Y.~Xu, M.~He, Y.~Wang, and W.~Liu,
  ``All you need is boundary: Toward arbitrary-shaped text spotting,'' in
  \emph{AAAI}, 2020.

\bibitem{xing2019convolutional}
X.~Linjie, T.~Zhi, H.~Weilin, and R.~S. Matthew, ``{Convolutional Character
  Networks},'' in \emph{ICCV}, 2019.

\bibitem{Liao2020MaskTV}
M.~Liao, G.~Pang, J.~Huang, T.~Hassner, and X.~Bai, ``Mask textspotter v3:
  Segmentation proposal network for robust scene text spotting,'' in
  \emph{ECCV}, 2020.

\bibitem{Liu2020ABCNetRS}
Y.~Liu, H.~Chen, C.~Shen, T.~He, L.~Jin, and L.~Wang, ``Abcnet: Real-time scene
  text spotting with adaptive bezier-curve network,'' in \emph{CVPR}, 2020, pp.
  9806--9815.

\bibitem{Fang2022ABINetAB}
S.~Fang, Z.~Mao, H.~Xie, Y.~Wang, C.~C. Yan, and Y.~Zhang, ``Abinet++:
  Autonomous, bidirectional and iterative language modeling for scene text
  spotting,'' \emph{TPAMI}, vol.~45, pp. 7123--7141, 2022.

\bibitem{Huang2022SwinTextSpotterST}
M.~Huang, Y.~Liu, Z.~Peng, C.~Liu, D.~Lin, S.~Zhu, N.~J. Yuan, K.~Ding, and
  L.~Jin, ``Swintextspotter: Scene text spotting via better synergy between
  text detection and text recognition,'' in \emph{CVPR}, 2022.

\bibitem{Carion2020EndtoEndOD}
N.~Carion, F.~Massa, G.~Synnaeve, N.~Usunier, A.~Kirillov, and S.~Zagoruyko,
  ``End-to-end object detection with transformers,'' in \emph{ECCV}, 2020.

\bibitem{Zhu2020DeformableDD}
X.~Zhu, W.~Su, L.~Lu, B.~Li, X.~Wang, and J.~Dai, ``Deformable detr: Deformable
  transformers for end-to-end object detection,'' in \emph{ICLR}, 2021.

\bibitem{Zhang2022TextST}
X.~Zhang, Y.~Su, S.~Tripathi, and Z.~Tu, ``Text spotting transformers,'' in
  \emph{CVPR}, 2022.

\bibitem{Kittenplon2022TowardsWT}
Y.~Kittenplon, I.~Lavi, S.~Fogel, Y.~Bar, R.~Manmatha, and P.~Perona, ``Towards
  weakly-supervised text spotting using a multi-task transformer,'' in
  \emph{CVPR}, 2022, pp. 4594--4603.

\bibitem{Peng2021SPTSST}
D.~Peng, X.~Wang, Y.~Liu, J.~Zhang, M.~Huang, S.~Lai, S.~Zhu, J.~Li, D.~Lin,
  C.~Shen, and L.~Jin, ``Spts: Single-point text spotting,'' in \emph{ACM MM},
  2021.

\bibitem{Ye2022DeepSoloLT}
M.~Ye, J.~Zhang, S.~Zhao, J.~Liu, T.~Liu, B.~Du, and D.~Tao, ``Deepsolo: Let
  transformer decoder with explicit points solo for text spotting,'' in
  \emph{CVPR}, 2023.

\bibitem{goh2021multimodal}
G.~Goh, N.~C. †, C.~V. †, S.~Carter, M.~Petrov, L.~Schubert, A.~Radford,
  and C.~Olah, ``Multimodal neurons in artificial neural networks,''
  \emph{Distill}, 2021, https://distill.pub/2021/multimodal-neurons.

\bibitem{Petroni2019LanguageMA}
F.~Petroni, T.~Rockt{\"a}schel, P.~Lewis, A.~Bakhtin, Y.~Wu, A.~H. Miller, and
  S.~Riedel, ``Language models as knowledge bases?'' in \emph{EMNLP}, 2019, p.
  1772–1791.

\bibitem{He2016DeepRL}
K.~He, X.~Zhang, S.~Ren, and J.~Sun, ``Deep residual learning for image
  recognition,'' in \emph{CVPR}, 2016, pp. 770--778.

\bibitem{Zhou2021LearningTP}
K.~Zhou, J.~Yang, C.~C. Loy, and Z.~Liu, ``Learning to prompt for
  vision-language models,'' \emph{IJCV}, p. 2337–2348, 2022.

\bibitem{Vaswani2017AttentionIA}
A.~Vaswani, N.~M. Shazeer, N.~Parmar, J.~Uszkoreit, L.~Jones, A.~N. Gomez,
  L.~Kaiser, and I.~Polosukhin, ``Attention is all you need,'' in
  \emph{NeurIPS}, 2017, pp. 1--11.

\bibitem{Karatzas2013ICDAR2R}
D.~Karatzas, F.~Shafait, S.~Uchida, M.~Iwamura, L.~G. i~Bigorda, S.~R. Mestre,
  J.~M. Romeu, D.~F. Mota, J.~Almaz{\'a}n, and L.-P. de~las Heras, ``Icdar 2013
  robust reading competition,'' in \emph{ICDAR}, 2013, pp. 1484--1493.

\bibitem{karatzas2015icdar}
D.~Karatzas, L.~Gomez-Bigorda \emph{et~al.}, ``{ICDAR 2015 competition on
  robust reading},'' in \emph{ICDAR}, 2015, pp. 1156--1160.

\bibitem{yao2012detecting}
C.~Yao, X.~Bai, W.~Liu, Y.~Ma, and Z.~Tu, ``Detecting texts of arbitrary
  orientations in natural images,'' in \emph{CVPR}.\hskip 1em plus 0.5em minus
  0.4em\relax IEEE, 2012, pp. 1083--1090.

\bibitem{liu2019curved}
Y.~Liu, L.~Jin, S.~Zhang, C.~Luo, and S.~Zhang, ``{Curved scene text detection
  via transverse and longitudinal sequence connection},'' \emph{PR}, vol.~90,
  pp. 337--345, 2019.

\bibitem{ch2019total}
C.-K. Ch’ng, C.~S. Chan, and C.-L. Liu, ``Total-text: toward orientation
  robustness in scene text detection,'' \emph{IJDAR}, pp. 1--22, 2019.

\bibitem{chng2019icdar2019}
C.-K. Chng, Y.~Liu, Y.~Sun, C.~C. Ng, C.~Luo, Z.~Ni, C.~Fang, S.~Zhang, J.~Han,
  E.~Ding \emph{et~al.}, ``{{ICDAR2019} Robust Reading Challenge on
  Arbitrary-Shaped Text (RRC-ArT)},'' in \emph{ICDAR}, 2019, pp. 1571--1576.

\bibitem{Nayef2017ICDAR2017RR}
N.~Nayef, F.~Yin, I.~Bizid, H.~Choi, Y.~Feng, D.~Karatzas, Z.~Luo, U.~Pal,
  C.~Rigaud, J.~Chazalon, W.~Khlif, M.~M. Luqman, J.-C. Burie, C.-L. Liu, and
  J.-M. Ogier, ``Icdar2017 robust reading challenge on multi-lingual scene text
  detection and script identification - rrc-mlt,'' in \emph{ICDAR}, vol.~01,
  2017, pp. 1454--1459.

\bibitem{nayef2019icdar2019}
N.~Nayef, Y.~Patel, M.~Busta, P.~N. Chowdhury, D.~Karatzas, W.~Khlif, J.~Matas,
  U.~Pal, J.-C. Burie, C.-l. Liu \emph{et~al.}, ``{{ICDAR2019} Robust Reading
  Challenge on Multi-lingual Scene Text Detection and
  Recognition--RRC-MLT-2019},'' in \emph{ICDAR}, 2019, pp. 1454--1459.

\bibitem{gupta2016synthetic}
A.~Gupta, A.~Vedaldi, and A.~Zisserman, ``Synthetic data for text localisation
  in natural images,'' in \emph{CVPR}, 2016, pp. 2315--2324.

\bibitem{Singh2021TextOCRTL}
A.~Singh, G.~Pang, M.~Toh, J.~Huang, W.~Galuba, and T.~Hassner, ``Textocr:
  Towards large-scale end-to-end reasoning for arbitrary-shaped scene text,''
  in \emph{CVPR}, 2021, pp. 8798--8808.

\bibitem{Zhan2019GADANGD}
F.~Zhan, C.~Xue, and S.~Lu, ``Ga-dan: Geometry-aware domain adaptation network
  for scene text detection and recognition,'' in \emph{ICCV}, 2019, pp.
  9104--9114.

\bibitem{Xu2019TextFieldLA}
Y.~Xu, Y.~Wang, W.~Zhou, Y.~Wang, Z.~Yang, and X.~Bai, ``Textfield: Learning a
  deep direction field for irregular scene text detection,'' \emph{TIP},
  vol.~28, pp. 5566--5579, 2019.

\bibitem{Baek2019CharacterRA}
Y.~Baek, B.~Lee, D.~Han, S.~Yun, and H.~Lee, ``Character region awareness for
  text detection,'' in \emph{CVPR}, 2019, pp. 9357--9366.

\bibitem{Wang2020ContourNetTA}
Y.~Wang, H.~Xie, Z.~Zha, M.~Xing, Z.~Fu, and Y.~Zhang, ``Contournet: Taking a
  further step toward accurate arbitrary-shaped scene text detection,'' in
  \emph{CVPR}, 2020, pp. 11\,750--11\,759.

\bibitem{Zhang2020DeepRR}
S.-X. Zhang, X.~Zhu, J.-B. Hou, C.~Liu, C.~Yang, H.~Wang, and X.-C. Yin, ``Deep
  relational reasoning graph network for arbitrary shape text detection,'' in
  \emph{CVPR}, 2020, pp. 9696--9705.

\bibitem{Raisi2021TransformerbasedTD}
Z.~Raisi, M.~A. Naiel, G.~Younes, S.~Wardell, and J.~S. Zelek,
  ``Transformer-based text detection in the wild,'' in \emph{IEEE CVPR
  Worksh.}, 2021, pp. 3156--3165.

\bibitem{Ye2020TextFuseNetST}
J.~Ye, Z.~Chen, J.~Liu, and B.~Du, ``Textfusenet: Scene text detection with
  richer fused features,'' in \emph{IJCAI}, 2020.

\bibitem{Liao2019RealTimeST}
M.~Liao, Z.~Wan, C.~Yao, K.~Chen, and X.~Bai, ``Real-time scene text detection
  with differentiable binarization and adaptive scale fusion,'' \emph{TPAMI},
  vol.~45, pp. 919--931, 2019.

\bibitem{Xing2019ConvolutionalCN}
L.~Xing, Z.~Tian, W.~Huang, and M.~R. Scott, ``Convolutional character
  networks,'' in \emph{ICCV}, 2019, pp. 9125--9135.

\bibitem{Baek2020CharacterRA}
Y.~Baek, S.~Shin, J.~Baek, S.~Park, J.~Lee, D.~Nam, and H.~Lee, ``Character
  region attention for text spotting,'' in \emph{ECCV}, 2020.

\bibitem{Qiao2020TextPT}
L.~Qiao, S.~Tang, Z.~Cheng, Y.~Xu, Y.~Niu, S.~Pu, and F.~Wu, ``Text perceptron:
  Towards end-to-end arbitrary-shaped text spotting,'' in \emph{AAAI}, 2020.

\bibitem{Feng2019TextDragonAE}
W.~Feng, W.~He, F.~Yin, X.-Y. Zhang, and C.-L. Liu, ``Textdragon: An end-to-end
  framework for arbitrary shaped text spotting,'' in \emph{ICCV}, 2019, pp.
  9075--9084.

\bibitem{Wang2021PANTE}
W.~Wang, E.~Xie, X.~Li, X.~Liu, D.~Liang, Y.~Zhibo, T.~Lu, and C.~Shen,
  ``Pan++: Towards efficient and accurate end-to-end spotting of
  arbitrarily-shaped text,'' \emph{TPAMI}, vol.~PP, 2021.

\bibitem{Wang2021PGNetRA}
P.~Wang, C.~Zhang, F.~Qi, S.~Liu, X.~Zhang, P.~Lyu, J.~Han, J.~Liu, E.~Ding,
  and G.~Shi, ``Pgnet: Real-time arbitrarily-shaped text spotting with point
  gathering network,'' in \emph{AAAI}, 2021.

\bibitem{Qiao2020MANGOAM}
L.~Qiao, Y.~Chen, Z.~Cheng, Y.~Xu, Y.~Niu, S.~Pu, and F.~Wu, ``Mango: A mask
  attention guided one-stage scene text spotter,'' in \emph{AAAI}, 2020.

\bibitem{Ronen2022GLASSGT}
R.~Ronen, S.~Tsiper, O.~Anschel, I.~Lavi, A.~Markovitz, and R.~Manmatha,
  ``Glass: Global to local attention for scene-text spotting,'' in \emph{ECCV},
  2022.

\bibitem{Liu2021ABCNetVA}
Y.~Liu, C.~Shen, L.~Jin, T.~He, P.~Chen, C.~Liu, and H.~Chen, ``Abcnet v2:
  Adaptive bezier-curve network for real-time end-to-end text spotting,''
  \emph{TPAMI}, vol.~44, pp. 8048--8064, 2021.

\bibitem{Wu2020SynthetictoRealUD}
W.~Wu, N.~Lu, E.~Xie, Y.~Wang, W.~Yu, C.~Yang, and H.~Zhou, ``Synthetic-to-real
  unsupervised domain adaptation for scene text detection in the wild,'' in
  \emph{ACCV}, 2020, pp. 1--14.

\bibitem{Kim2016PVANETDB}
K.-H. Kim, Y.~Cheon, S.~Hong, B.-S. Roh, and M.~Park, ``Pvanet: Deep but
  lightweight neural networks for real-time object detection,'' \emph{arXiv:
  Comp. Res. Repository}, 2016.

\bibitem{Liao2020SynthText3DSS}
M.~Liao, B.~Song, M.~He, S.~Long, C.~Yao, and X.~Bai, ``Synthtext3d:
  synthesizing scene text images from 3d virtual worlds,'' \emph{Sci.
  China-Inf. Sci.}, vol.~63, pp. 1--14, 2020.

\bibitem{Zhu2017UnpairedIT}
J.-Y. Zhu, T.~Park, P.~Isola, and A.~A. Efros, ``Unpaired image-to-image
  translation using cycle-consistent adversarial networks,'' in \emph{ICCV},
  2017, pp. 2242--2251.

\bibitem{Lin2018STGANST}
C.-H. Lin, E.~Yumer, O.~Wang, E.~Shechtman, and S.~Lucey, ``St-gan: Spatial
  transformer generative adversarial networks for image compositing,'' in
  \emph{CVPR}, 2018, pp. 9455--9464.

\bibitem{PolarDet}
P.~Zhao, Z.~Qu, Y.~Bu, W.~Tan, and Q.~Guan, ``Polardet: A fast, more precise
  detector for rotated target in aerial images,'' \emph{Int. J. Remote Sens.},
  vol.~42, no.~15, pp. 5821--5851, 2021.

\bibitem{RDD}
B.~Zhong and K.~Ao, ``Single-stage rotation-decoupled detector for oriented
  object,'' \emph{Remote Sens.}, vol.~12, no.~19, p. 3262, 2020.

\bibitem{GWD}
X.~Yang, J.~Yan, Q.~Ming, W.~Wang, X.~Zhang, and Q.~Tian, ``Rethinking rotated
  object detection with gaussian wasserstein distance loss,'' in \emph{ICML},
  2021.

\bibitem{KLD}
X.~Yang, X.~Yang, J.~Yang, Q.~Ming, W.~Wang, Q.~Tian, and J.~Yan, ``Learning
  high-precision bounding box for rotated object detection via kullback-leibler
  divergence,'' in \emph{NeurIPS}, 2021.

\bibitem{lin2017focal}
T.-Y. Lin, P.~Goyal, R.~Girshick, K.~He, and P.~Doll{\'a}r, ``Focal loss for
  dense object detection,'' in \emph{ICCV}, 2017.

\bibitem{tian2019fcos}
Z.~Tian, C.~Shen, H.~Chen, and T.~He, ``Fcos: Fully convolutional one-stage
  object detection,'' in \emph{ICCV}, 2019.

\bibitem{zhang2020bridging}
S.~Zhang, C.~Chi, Y.~Yao, Z.~Lei, and S.~Z. Li, ``Bridging the gap between
  anchor-based and anchor-free detection via adaptive training sample
  selection,'' in \emph{CVPR}, 2020, pp. 9759--9768.

\bibitem{Xia2017DOTAAL}
G.-S. Xia, X.~Bai, J.~Ding, Z.~Zhu, S.~J. Belongie, J.~Luo, M.~Datcu,
  M.~Pelillo, and L.~Zhang, ``Dota: A large-scale dataset for object detection
  in aerial images,'' in \emph{CVPR}, 2017, pp. 3974--3983.

\bibitem{Yao2014AUF}
C.~Yao, X.~Bai, and W.~Liu, ``A unified framework for multi-oriented text
  detection and recognition,'' \emph{TIP}, vol.~23, pp. 4737--4749, 2014.

\end{thebibliography}
